\documentclass[10pt,twocolumn,letterpaper]{article}

\usepackage[pagenumbers]{cvpr} 
\usepackage[accsupp]{axessibility}
\usepackage{xspace}
\usepackage[dvipsnames]{xcolor}

\usepackage{graphicx}
\usepackage{amsmath}
\usepackage{amssymb}
\usepackage{booktabs}
\usepackage{microtype}
\usepackage{pifont}
\usepackage{color}
\usepackage{placeins}
\usepackage{float}
\usepackage{soul}
\usepackage{enumitem}
\usepackage{adjustbox}
\usepackage{commath}
\usepackage{bbm}
\usepackage{multirow}
\usepackage{caption}
\usepackage{longtable}
\usepackage{array}
\usepackage{xcolor}
\usepackage{svg}
\usepackage{makecell}
\usepackage{tcolorbox}
\usepackage{fontawesome}
\usepackage{inconsolata}
\usepackage{algorithm}
\usepackage{algorithmic}
\usepackage{afterpage}
\usepackage{amsfonts}
\usepackage{gensymb}

\usepackage[dvipsnames]{xcolor}

\newcommand{\holodecklogo}[1][1.85em]{%
  \raisebox{-0.3\height}{\includegraphics[height=#1]{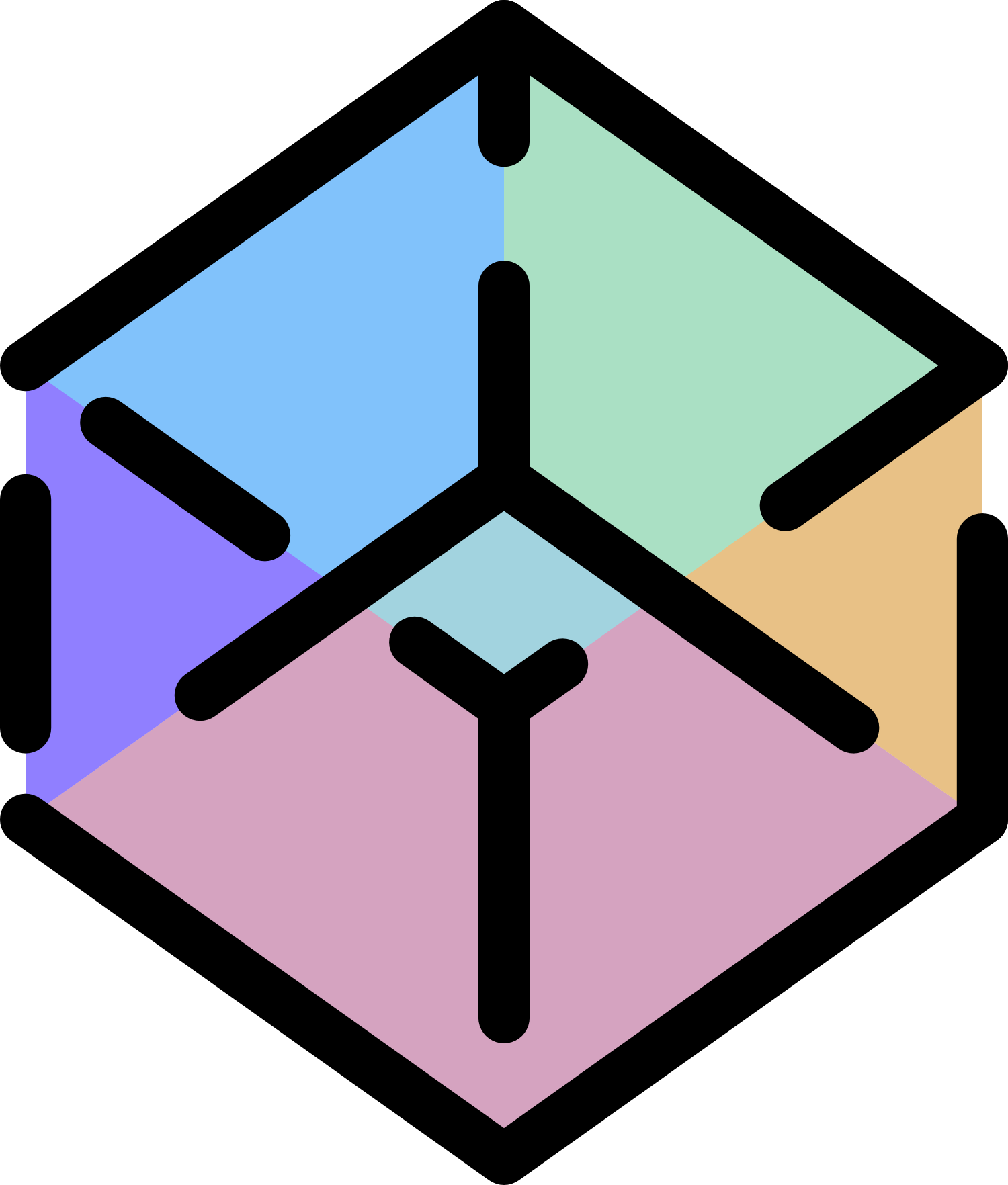}}
}

\newcommand{\holodeck}{\textsc{Holodeck}\xspace}
\newcommand{\procthor}{\textsc{ProcTHOR}\xspace}
\newcommand{\noveltythor}{\textsc{NoveltyTHOR}\xspace}

\definecolor{cvprblue}{rgb}{0.21,0.49,0.74}
\usepackage[pagebackref,breaklinks,colorlinks,citecolor=cvprblue]{hyperref}

\def\paperID{595} 
\def\confName{CVPR}
\def\confYear{2024}
\title{\holodecklogo\holodeck: Language Guided Generation of 3D Embodied AI Environments}

\author{
Yue Yang*$^{1}$, Fan-Yun Sun*$^{2}$, Luca Weihs*$^{4}$, Eli Vanderbilt${^4}$, Alvaro Herrasti${^4}$, \\
Winson Han$^4$,  Jiajun Wu$^{2}$, Nick Haber$^2$, Ranjay Krishna$^{3,4}$, Lingjie Liu${^1}$, \\
Chris Callison-Burch${^1}$, Mark Yatskar${^1}$, Aniruddha Kembhavi$^{3,4}$, Christopher Clark$^4$ \vspace{0.1cm} \\ 
$^1$University of Pennsylvania, $^2$Stanford University, \\
$^3$University of Washington, $^4$Allen Institute for Artificial Intelligence \vspace{0.1cm} \\
{\tt \href{https://yueyang1996.github.io/holodeck/}{yueyang1996.github.io/holodeck/}} \vspace{-0.2cm}
}

\begin{document}
\twocolumn[{%
\maketitle
\begin{figure}[H]
\hsize=\textwidth
\centering
\includegraphics[width=16cm]{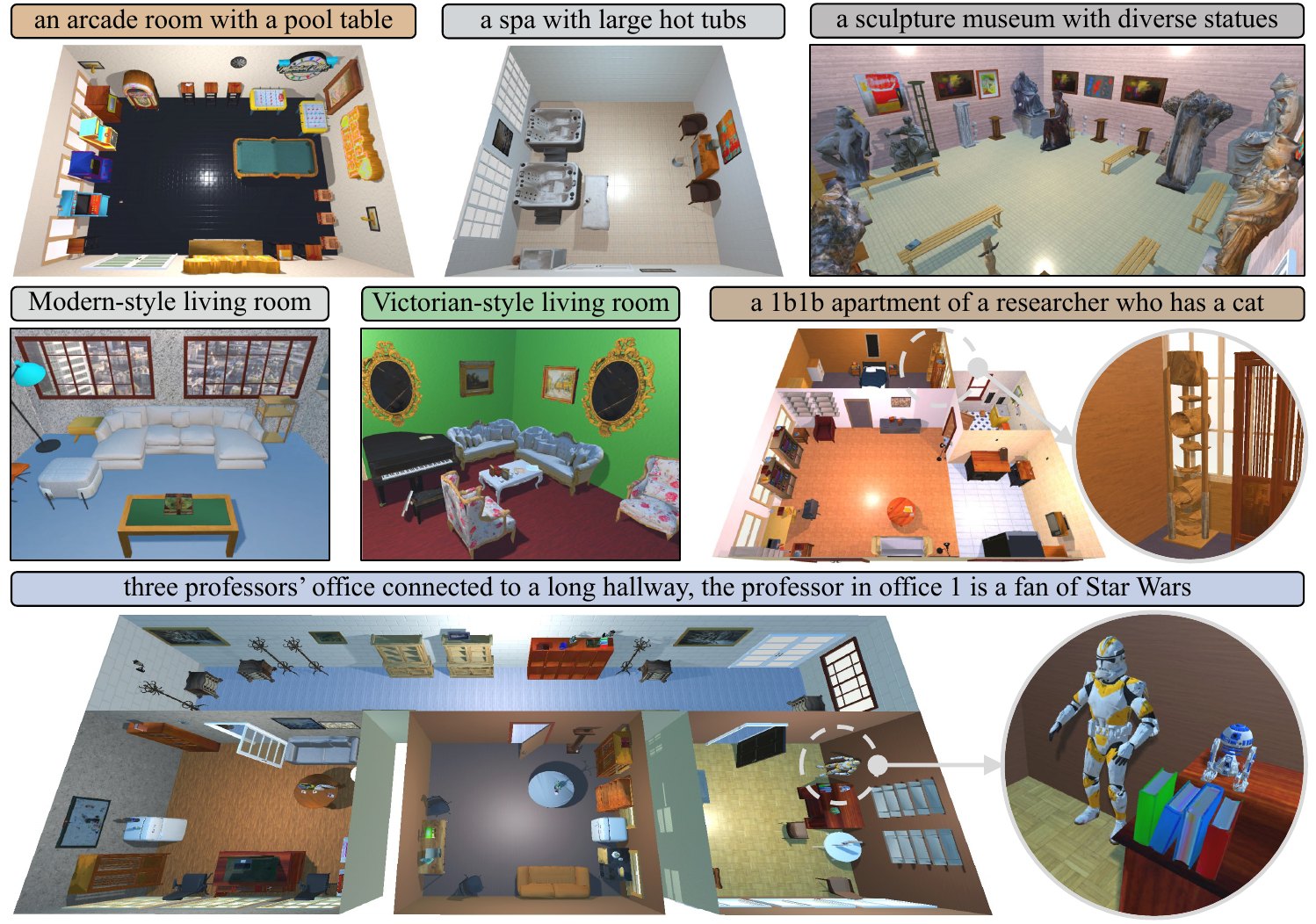}
    \caption{Example outputs of \holodeck---a large language model powered system, which can generate diverse types of environments (arcade, spa, museum), customize for styles (Victorian-style), and understand fine-grained requirements (``has a cat'', ``fan of Star Wars'').}
    \label{fig: demo examples}
\end{figure}
\vspace{-0.1cm}
}]


\begin{abstract}
3D simulated environments play a critical role in Embodied AI, but their creation requires expertise and extensive manual effort, restricting their diversity and scope.\renewcommand{\thefootnote}{\fnsymbol{footnote}}\footnotetext[1]{Equal technical contribution. Work done while at PRIOR@AI2.}\renewcommand{\thefootnote}{\arabic{footnote}}\setcounter{footnote}{0}
To mitigate this limitation, we present \holodeck, a system that generates 3D environments to match a user-supplied prompt fully automatedly.
\holodeck can generate diverse scenes, e.g., arcades, spas, and museums, adjust the designs for styles, and can capture the semantics of complex queries such as ``apartment for a researcher with a cat'' and ``office of a professor who is a fan of Star Wars''.
\holodeck leverages a large language model (i.e., GPT-4) for common sense knowledge about what the scene might look like and uses a large collection of 3D assets from Objaverse to populate the scene with diverse objects.
To address the challenge of positioning objects correctly, we prompt GPT-4 to generate spatial relational constraints between objects and then optimize the layout to satisfy those constraints.
Our large-scale human evaluation shows that annotators prefer \holodeck over manually designed procedural baselines in residential scenes and that \holodeck can produce high-quality outputs for diverse scene types.
We also demonstrate an exciting application of \holodeck in Embodied AI, training agents to navigate in novel scenes like music rooms and daycares without human-constructed data, which is a significant step forward in developing general-purpose embodied agents. 
\end{abstract}

\section{Introduction}
The predominant approach in training embodied agents involves learning in simulators~\cite{ai2thor, xia2018gibson, puig2018virtualhome, savva2019habitat, phone2proc, Kadian2019Sim2RealPD}. Generating realistic, diverse, and interactive 3D environments plays a crucial role in the success of this process.

Existing Embodied AI environments are typically crafted through manual design \cite{ai2thor, deitke2020robothor,  li2023behavior, gan2020threedworld}, 3D scanning \cite{savva2019habitat, ramakrishnan2021hm3d, phone2proc}, or procedurally generated with hard-coded rules \cite{procthor}. However, these methods require considerable human effort that involves designing a complex layout, using assets supported by an interactive simulator, and placing them into scenes while ensuring semantic consistency between the different scene elements. Therefore, prior work on producing 3D environments mainly focuses on limited environment types.
To move beyond these limitations, recent works adapt 2D foundational models to generate 3D scenes from text~\cite{zhang2023text2nerf,hollein2023text2room,fridman2023scenescape}. However, these models often produce scenes with significant artifacts, such as mesh distortions, and lack the interactivity necessary for Embodied AI.
Moreover, there are models tailored for specific tasks like floor plan generation~\cite{hu2020graph2plan,shabani2023housediffusion} or object arrangement~\cite{wei2023lego,paschalidou2021atiss}. Although effective in their respective domains, they lack overall scene consistency and rely heavily on task-specific datasets.

\begin{figure*}[!t]
\hsize=\textwidth
\centering
\includegraphics[width=0.96\textwidth]{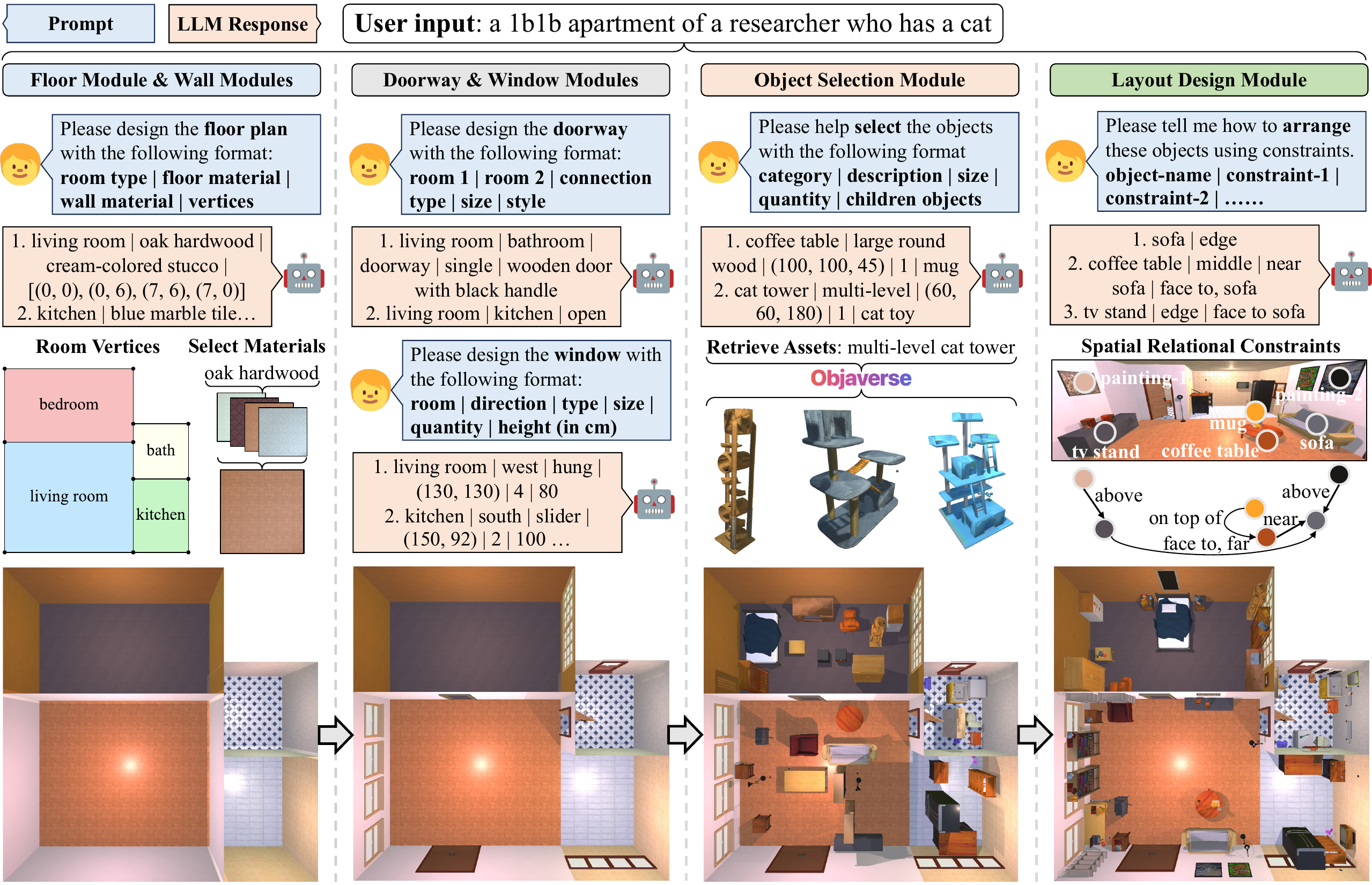}
    \caption{Given a text input, \holodeck generates the 3D environment through multiple rounds of conversation with an LLM.}
    \label{fig: pipeline}
    \vspace{-0.5cm}
\end{figure*}

In light of these challenges, we present \textbf{\holodeck}, a language-guided system built upon AI2-THOR \cite{ai2thor}, to automatically generate diverse, customized, and interactive 3D embodied environments from textual descriptions. Shown in Figure \ref{fig: pipeline}, given a description (e.g., \emph{a 1b1b apartment of a researcher who has a cat}), \holodeck uses a Large Language Model (GPT-4 \cite{openai2023gpt}) to design the floor plan, assign suitable materials, install the doorways and windows and arrange 3D assets coherently in the scene using constraint-based optimization. \holodeck chooses from over 50K diverse and high-quality 3D assets from Objaverse~\cite{deitke2023objaverse} to satisfy a myriad of environment descriptions.
 

Motivated by the emergent abilities of Large Language Models (LLMs) \cite{wei2022emergent}, \holodeck exploits the commonsense priors and spatial knowledge inherently present in LLMs. This is exemplified in Figure \ref{fig: demo examples}, where \holodeck creates diverse scene types such as \textit{arcade}, \textit{spa} and \textit{museum}, interprets specific and abstract prompts by placing relevant objects appropriately into the scene, e.g., an ``R2-D2''\footnote{A fictional robot character in the Star Wars.} on the desk for ``a fan of Star Wars". Beyond object selection and layout design, \holodeck showcases its versatility in style customization, such as creating a scene in a ``Victorian-style" by applying appropriate textures and designs to the scene and its objects. Moreover, \holodeck demonstrates its proficiency in spatial reasoning, like devising floor plans for ``three professors' offices connected by a long hallway'' and having regular arrangements of objects in the scenes.
Overall, \holodeck offers a broad coverage approach to 3D environment generation, where textual prompts unlock new levels of control and flexibility in scene creation.

The effectiveness of \holodeck is assessed through its scene generation quality and applicability to Embodied AI. Through large-scale user studies involving 680 participants, we demonstrate that \holodeck significantly surpasses existing procedural baseline \procthor \cite{procthor} in generating residential scenes and achieves high-quality outputs for various scene types. For the Embodied AI experiments, we focus on \holodeck's application in aiding zero-shot object navigation in previously unseen scene types. 
We show that agents trained on scenes generated by \holodeck can navigate better in novel environments (e.g., \textit{Daycare} and \textit{Gym}) designed by experts.


To summarize, our contributions are three-fold: (1) We propose \holodeck, a language-guided system capable of generating diverse, customized, and interactive 3D environments based on textual descriptions; (2) The human evaluation validates \holodeck's capability of generating residential and diverse scenes with accurate asset selection and realistic layout design; (3) Our experiments demonstrate that \holodeck can aid Embodied AI agents in adapting to new scene types and objects during object navigation tasks.




\section{Related Work}
\medbreak
\noindent \textbf{Embodied AI Environments.} Previous work mainly relies on 3D artists to design the environments \cite{ai2thor, li2023behavior, puig2018virtualhome, xia2018gibson, deitke2020robothor, gan2020threedworld, khanna2023hssd}, which is hard to scale up or construct scenes from 3D scans \cite{savva2019habitat, szot2021habitat, ramakrishnan2021hm3d} to reduce human labor, but scenes are less interactive.
The procedural generation framework \procthor~\cite{procthor} showcases its potential to generate large-scale interactive environments for training embodied agents. Phone2Proc \cite{phone2proc} uses a phone scan to create training scenes that are semantically similar to the desired real-world scene. A concurrent work, RoboGen~\cite{wang2023robogen}, proposes to train robots by generating diversified tasks and scenes. These works parallel our concept, \holodeck, which aims to train generalizable embodied agents and presents an avenue for further exploration in text-driven 3D interactive scene generation.

\smallbreak
\noindent \textbf{Large Language Model for Scene Design.}
Many works on scene design either learn spatial knowledge priors from existing 3D scene databases~\cite{chang2017sceneseer,tan2019text2scene,ma2018language,tang2023diffuscene,zhao2023roomdesigner,wang2021sceneformer,wei2023lego} or leverage user input and refine the 3D scene iteratively~\cite{chang2014interactive,cheng2019interactive}. However, having to learn from datasets of limited categories such as 3D-FRONT~\cite{fu20213d} restricts their applicability. Recently, Large Language Models (LLMs) were shown to be useful in generating 3D scene layouts~\cite{feng2023layoutgpt,lin2023towards}.
However, their methods of having LLMs directly output numerical values can yield layouts that defy physical plausibility (e.g., overlapping assets). In contrast, 
\holodeck uses LLMs to sample spatial relational constraints and a solver to optimize the layout, ensuring physically plausible scene arrangements. 
Our human study shows a preference for \holodeck-generated layouts over those generated end-to-end by LLMs. (see Sec~\ref{sec: layout ablation}).

\smallbreak
\noindent\textbf{Text-driven 3D Generation.} Early endeavors in 3D generation focus on learning the distribution of 3D shapes and/or textures from category-specific datasets~\cite{wu2016learning,yang2019pointflow,zhou20213d,henzler2019escaping,nguyen2019hologan}. 
Subsequently, the advent of large vision-language models like CLIP~\cite{radford2021learning} enables zero-shot generation of 3D textures and objects \cite{huang2023aladdin,mildenhall2021nerf,poole2022dreamfusion,metzer2023latent,lin2023magic3d,gu2023nerfdiff}.
These works excel at generating 3D objects but struggle to generate complex 3D scenes. 
More recently, emerging works generate 3D scenes by combining pre-trained text-to-image models with depth prediction 
algorithms to produce either textured meshes or NeRFs~\cite{fridman2023scenescape,hollein2023text2room,zhang2023text2nerf}. However, these approaches yield 3D representations that lack modular composability and interactive affordances, limiting their use in embodied AI. In contrast, \holodeck utilizes a comprehensive 3D asset database to generate semantically precise, spatially efficient, and interactive 3D environments suitable for training embodied agents.

\section{\holodeck}
\begin{figure*}[!t]
\centering
\includegraphics[width=\textwidth]{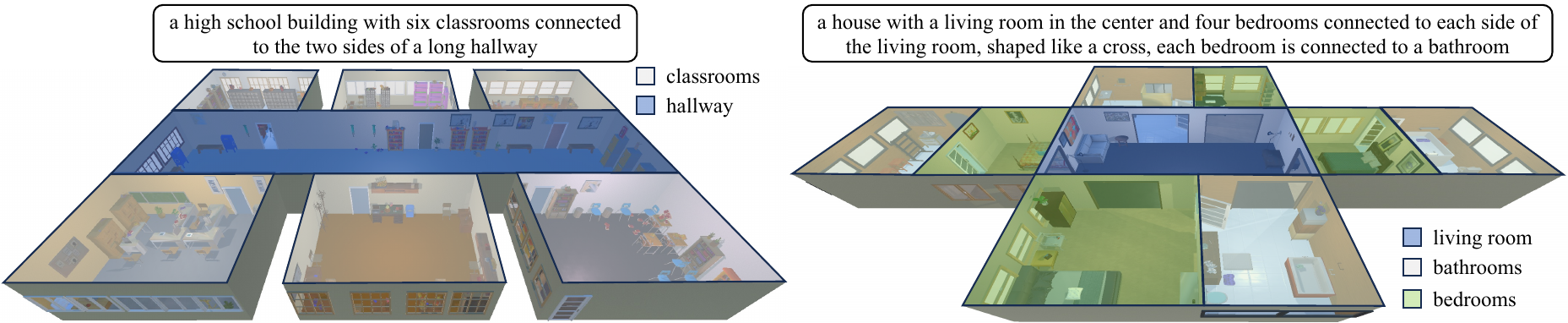}
    \caption{\textbf{Floorplan Customizability.} \holodeck can interpret complicated input and craft reasonable floor plans correspondingly.}
    \label{fig: floor_plan}
    \vspace{-0.2cm}
\end{figure*}

\begin{figure*}[!t]
\centering
\includegraphics[width=\textwidth]{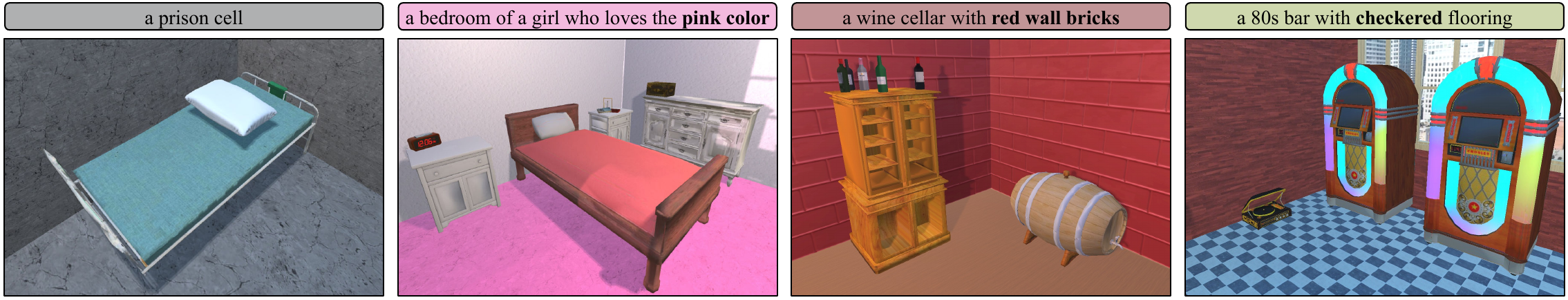}
    \caption{\textbf{Material Customizability.} \holodeck can select appropriate floor and wall materials to make the scenes more realistic.}
    \label{fig: material}
    \vspace{-0.2cm}
\end{figure*}

\begin{figure*}[!t]
\centering
\includegraphics[width=\textwidth]{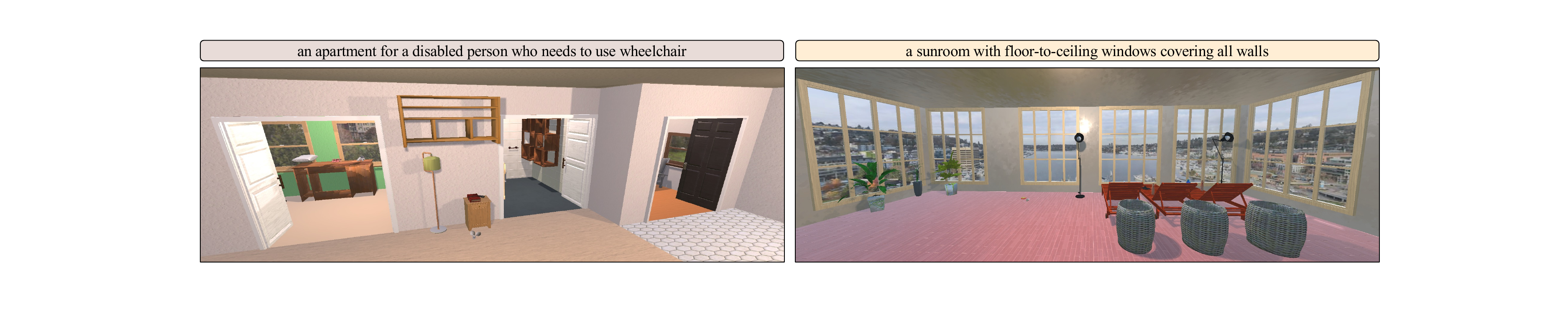}
    \caption{\textbf{Door \& window Customizability.} \holodeck can adjust the size, quantity, position, etc., of doors \& windows based on the input.}
    \label{fig: window_door}
    \vspace{-0.35cm}
\end{figure*}
\holodeck is a promptable system based on AI2-THOR \cite{ai2thor, procthor}, enriched with massive assets from Objaverse~\cite{deitke2023objaverse}, which can produce diverse, customized, and interactive Embodied AI environments with the guidance of large language models.
 
As shown in Figure \ref{fig: pipeline}, \holodeck employs a systematic approach to scene construction, utilizing a series of specialized modules: (1) the \textit{Floor \& Wall Module} develop floor plans, constructs wall structures and selects appropriate materials for the floors and walls; (2) the \textit{Doorway \& Window Module} integrates doorways and windows into the environment; (3) the \textit{Object Selection Module} retrieves appropriate 3D assets from Objaverse, and (4) the \textit{Constraint-based Layout Design Module} arranges the assets within the scene by utilizing spatial relational constraints to ensure that the layout of objects is realistic.


In the following sections, we introduce our prompting approach that converts high-level user natural language specifications into a series of language model queries for constructing layouts. 
We then provide a detailed overview of each module shown in Figure~\ref{fig: pipeline} and how they contribute to the final scene. 
Finally, we illustrate how \holodeck leverages Objaverse assets to ensure diversity in scene creation and efficiency for Embodied AI applications. Comprehensive details of \holodeck can be found in the supplement.


\smallbreak
\noindent \textbf{Overall Prompt Design.} 
Each module in Figure~\ref{fig: pipeline} takes information from a language model and converts it to elements included in the final layout.
An LLM prompt is designed for each module with three elements: (1) \textit{Task Description}: outlines the context and goals of the task; (2) \textit{Output Format}: specifies the expected structure and type of outputs and (3) \textit{One-shot Example}: a concrete example to assist the LLM's comprehension of the task. The text within the blue dialog boxes of Figure \ref{fig: pipeline} represents examples of simplified prompts\footnote{The complete prompts (available in the supplementary materials) include additional guidance for LLMs to avoid common errors we observe. For example, by adding a sentence, ``the minimal area per room is 9 m$^2$'', \holodeck can avoid generating overly small rooms.}. LLM's high-level responses to these prompts are post-processed and then used as input arguments for the modules to yield low-level specifications of the scene.

\smallbreak
\noindent The \textbf{Floor \& Wall Module}, illustrated in the first panel of Figure~\ref{fig: pipeline}, is responsible for creating floor plans, constructing wall structures, and selecting materials for floors and walls.
Each room is represented as a rectangle, defined by four tuples that specify the coordinates of its corners.
GPT-4 directly yields the coordinates for placing the rooms and suggests realistic dimensions and connectivity for these rooms.
Figure \ref{fig: floor_plan} illustrates several examples of diverse layouts this module proposes where \holodeck generates prompt-appropriate, intricate, multi-room floor plans. 
\begin{figure*}[!t]
\centering
\includegraphics[width=\textwidth]{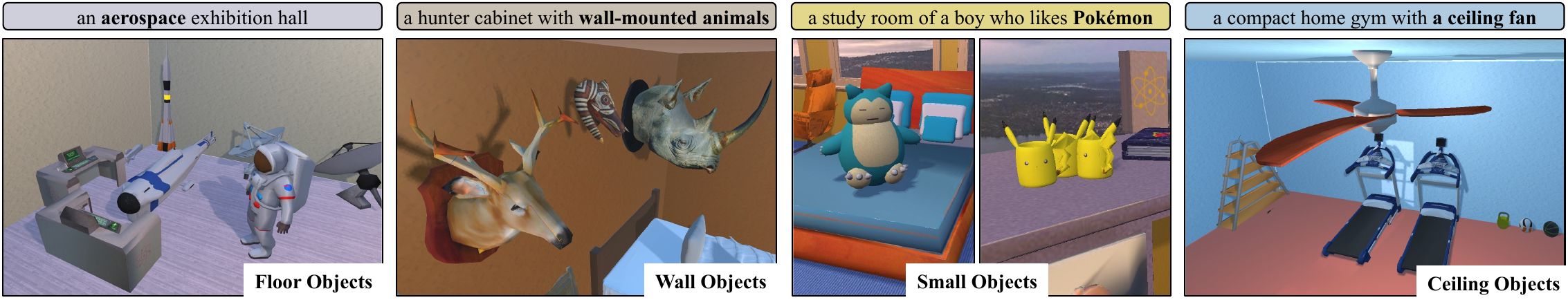}
    \caption{\textbf{Objects Customizability.} \holodeck can select and place appropriate floor/wall/small/ceiling objects conditioned on the input.}
    \label{fig: asset}
    \vspace{-0.2cm}
\end{figure*}

\begin{figure*}[!t]
\centering
\includegraphics[width=\textwidth]{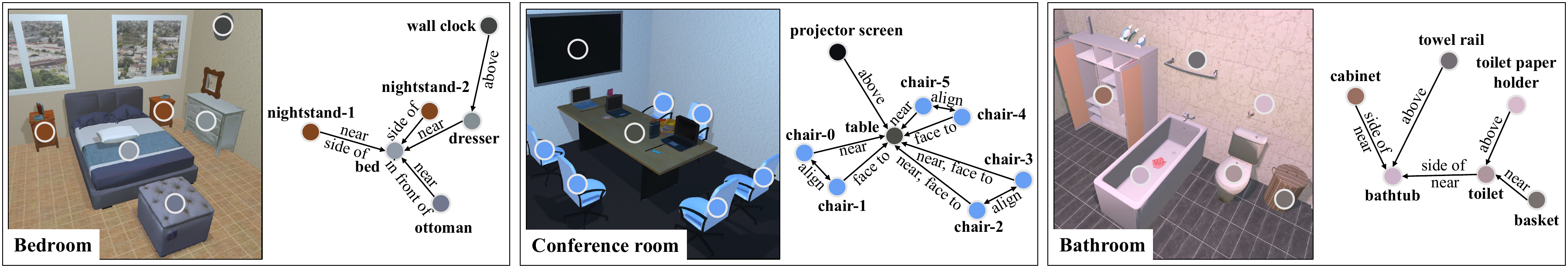}
    \caption{Examples of \textbf{Spatial Relational Constraints} generated by LLM and their solutions found by our constraint satisfaction algorithm.}
    \label{fig: constraints}
    \vspace{-0.2cm}
\end{figure*}

\begin{figure*}[!t]
\centering
\includegraphics[width=\textwidth]{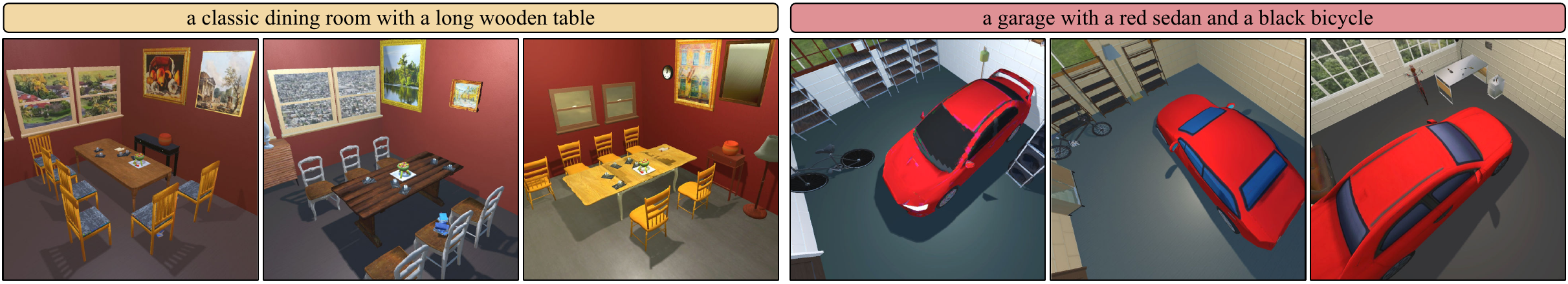}
    \caption{\textbf{Output Diversity.} \holodeck can generate \textbf{multiple variants} for the same input with different assets and layouts.}
    \label{fig: variants}
    \vspace{-0.5cm}
\end{figure*}
\smallbreak
This module also chooses materials for the floors and walls, which is crucial for enhancing the realism of environments.
\holodeck can match LLM proposals to one of 236 materials, each available in 148 colors, enabling semantic customization of scenes. 
As shown in Figure \ref{fig: material}, \holodeck can generate scenes with suitable materials based on the type of scene, such as opting for concrete walls and floors in a \textit{prison cell} scenario.
Inputs with specific texture requirements are often reflected in the final design, for example, ``pink color", ``red wall bricks," and ``checkered floor".
\smallbreak
\noindent The \textbf{Doorway \& Window Module}, illustrated in the second panel of Figure~\ref{fig: pipeline}, is responsible for proposing room connections and windows. 
Each of these two properties is queried separately from the LLM.
The LLM can propose doorways and windows that match 40 door styles and 21 window types, each of which can be modified by several properties, including size, height, quantity, etc.
For instance, Figure \ref{fig: window_door} shows \holodeck's tailored designs on doors and windows, such as wider doors for ``wheelchair accessibility'' and multiple floor-to-ceiling windows in a ``sunroom'' setting.
\smallbreak
\noindent The \textbf{Object Selection Module}, illustrated in the third panel of Figure~\ref{fig: pipeline}, allows the LLM to propose objects that should be included in the layout.
Leveraging the extensive Objaverse asset collection, \holodeck can fetch and place diverse objects in the scene.
Queries are constructed with LLM-proposed descriptions and dimensions, like ``multi-level cat tower, 60 $\times$ 60 $\times$ 180 (cm)'' to retrieve the optimal asset from Objaverse.
The retrieval function\footnote{We use CLIP \cite{radford2021learning} to measure the visual similarity, Sentence-BERT \cite{reimers-2019-sentence-bert} for the textual similarity, and 3D bounding box sizes for the dimension.} considers visual and textual similarity and dimensions to ensure the assets match the design. Figure \ref{fig: asset} shows the capability of \holodeck to customize diverse objects on the floor, walls, on top of other items, and even on the ceiling.
\smallbreak
\noindent The \textbf{Constraint-based Layout Design Module}, illustrated in the fourth panel of Figure~\ref{fig: pipeline}, generates the positioning and orientation of objects. Previous work \cite{feng2023layoutgpt} shows LLM can directly provide the absolute value of the object's bounding box. However, when attempting to place a diverse lot of assets within environments, this method frequently leads to out-of-boundary errors and object collisions. To address this, instead of letting LLM directly operate on numerical values, we propose a novel constraint-based approach that employs LLM to generate spatial relations between the objects, e.g., ``coffee table, in front of, sofa'', and optimize the layout based on the constraints. Given the probabilistic nature of LLMs, \holodeck can yield multiple valid layouts given the same prompt as shown in Figure \ref{fig: variants}.
\smallbreak
\noindent \textit{Spatial Relational Constraints.} We predefined ten types of constraints, organized into five categories: (1) Global: \textit{edge}, \textit{middle}; (2) Distance: \textit{near}, \textit{far}; (3) Position: \textit{in front of}, \textit{side of}, \textit{above}, \textit{on top of}; (4) Alignment: \textit{center aligned} and (5) Rotation: \textit{face to}. LLM selects a subset of constraints for each object, forming a scene graph for the room (examples shown in Figure \ref{fig: constraints}). Those constraints are treated softly, allowing for certain violations when finding a layout to satisfy all constraints is not feasible. Besides those soft constraints, we enforce hard constraints to prevent object collisions and ensure that all objects are within the room's boundaries. 
\smallbreak
\noindent \textit{Constraint Satisfaction.} We first reformulate the spatial relational constraints defined above into mathematical conditions (e.g., two objects are center-aligned if they share the same $x$ or $y$ coordinate). To find layouts that satisfy constraints sampled by LLMs, we adopt an optimization algorithm to place objects autoregressively. The algorithm first uses LLM to identify an anchor object and then explores placements for the anchor object. Subsequently, it employs Depth-First-Search (DFS)\footnote{Given the linear nature of constraints, a Mixed Integer Linear Programming (MILP) solver can also be employed. While we assume the DFS solver in our experiments, we analyze the MILP solver in the supplements.} to find valid placements for the remaining objects. A placement is only valid if all the hard constraints are satisfied.
For example, in Figure \ref{fig: constraints}, \textit{bed} is selected as the anchor object in the \textit{bedroom}, and the \textit{nightstands} are placed subsequently. The algorithm is executed for a fixed time (30 seconds) to get multiple candidate layouts and return the one that satisfies the most total constraints. We verify the effectiveness of our constraint-based layout in Sec \ref{sec: layout ablation}.


\smallbreak
\noindent \textbf{Leveraging Objaverse Assets}, \holodeck is able to support the creation of diverse and customized scenes. 
We curate a subset of assets suitable for indoor design from \href{https://objaverse.allenai.org/objaverse-1.0/}{Objaverse 1.0}. These assets are further annotated by GPT-4-Vison \cite{openai_2023_gpt4vision} automatically with additional details, including textual descriptions, scale, canonical views, etc.\footnote{GPT-4-Vision can take in multiple images, we prompt it with multi-view screenshots of 3D assets to get the annotations.} Together with the assets from \procthor, our library encompasses 51,464 annotated assets.
To import Objaverse assets into AI2-THOR for embodied AI applications, we optimize the assets by reducing mesh counts to minimize the loading time in AI2-THOR, generating visibility points and colliders. More details on importing Objaverse assets into AI2-THOR are available in the supplementary materials.

In the following sections, we will evaluate the quality and utility of the scenes generated by \holodeck.

\section{Human Evaluation}
\begin{figure}[!t]
\centering
\includegraphics[width=8.3cm]{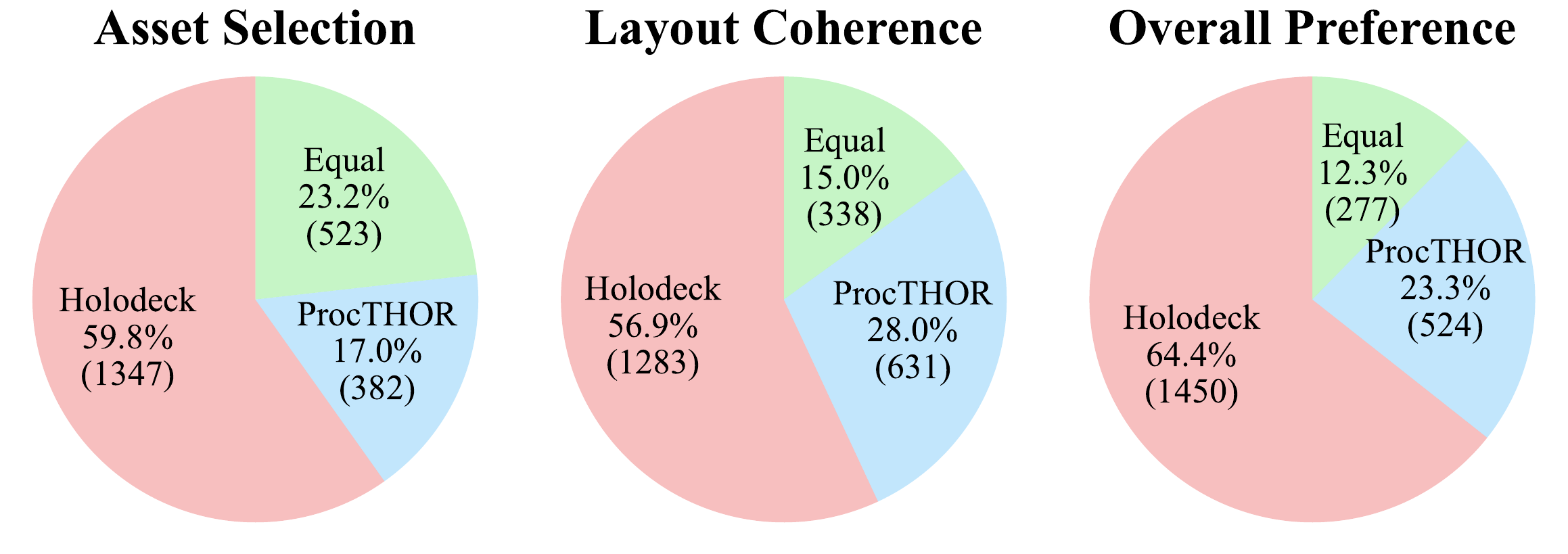}
    \caption{Comparative human evaluation of \holodeck and \procthor across three criteria. The pie charts show the distribution of annotator preferences, showing both the percentage and the actual number of annotations favoring each system.}
    \label{fig: holodeck vs. procthor}
    \vspace{-0.5cm}
\end{figure}

We conduct comprehensive human evaluations to assess the quality of \holodeck scenes, with a total of 680 graduate students participating in three user studies: (1) a comparative analysis on \textbf{residential scenes} with \procthor as the baseline; (2) an examination of \holodeck's ability in generating \textbf{diverse scenes}, and (3) an ablation study to validate the effectiveness of our \textbf{layout design} method. Through these user studies, we demonstrate that \holodeck can create residential scenes of better quality than previous work while being able to extend to a wider diversity of scene types.

\subsection{Comparative Analysis on Residential Scenes} \label{sec: holodeck vs procthor}
This study collects human preference scores to compare \holodeck with \procthor \cite{procthor}, the sole prior work capable of generating complete, interactable scenes. Our comparison focuses on residential scenes, as \procthor is limited to four types: \textit{bathroom}, \textit{bedroom}, \textit{kitchen}, and \textit{living room}.


\smallbreak
\noindent \textbf{Setup.} We prepared 120 scenes for human evaluation, comprising 30 scenes per scene type, for both \holodeck and the \procthor baseline. The \procthor baseline has access to the same set of Objaverse assets as \holodeck. For \holodeck, we take the scene type, e.g., ``bedroom'', as the prompt to generate the scenes. We pair scenes of the same scene type from the two systems, resulting in 120 paired scenes for human evaluation. For each paired scene, we display two shuffled top-down view images of the scenes from the two systems.
We ask the annotator to choose which scene is better or equally good based on three questions: (1) \textbf{Asset Selection}: which selection of 3D assets is more accurate/faithful to the scene type? (2) \textbf{Layout Coherence}: which arrangement of 3D assets adheres better to realism and common sense (considering the position and orientation)? and (3) \textbf{Overall Preference}: which of the two scenes would you prefer given the scene type? 
\smallbreak
\noindent \textbf{Humans prefer \holodeck over \procthor.}
Figure \ref{fig: holodeck vs. procthor} presents a clear preference for \holodeck in the comparative human evaluation against \procthor, with a majority of annotators favoring \holodeck for Asset Selection (59.8\%), Layout Coherence (56.9\%), and showing a significant preference in Overall Preference (64.4\%).
\begin{figure}[!t]
\centering
\includegraphics[width=8.3cm]{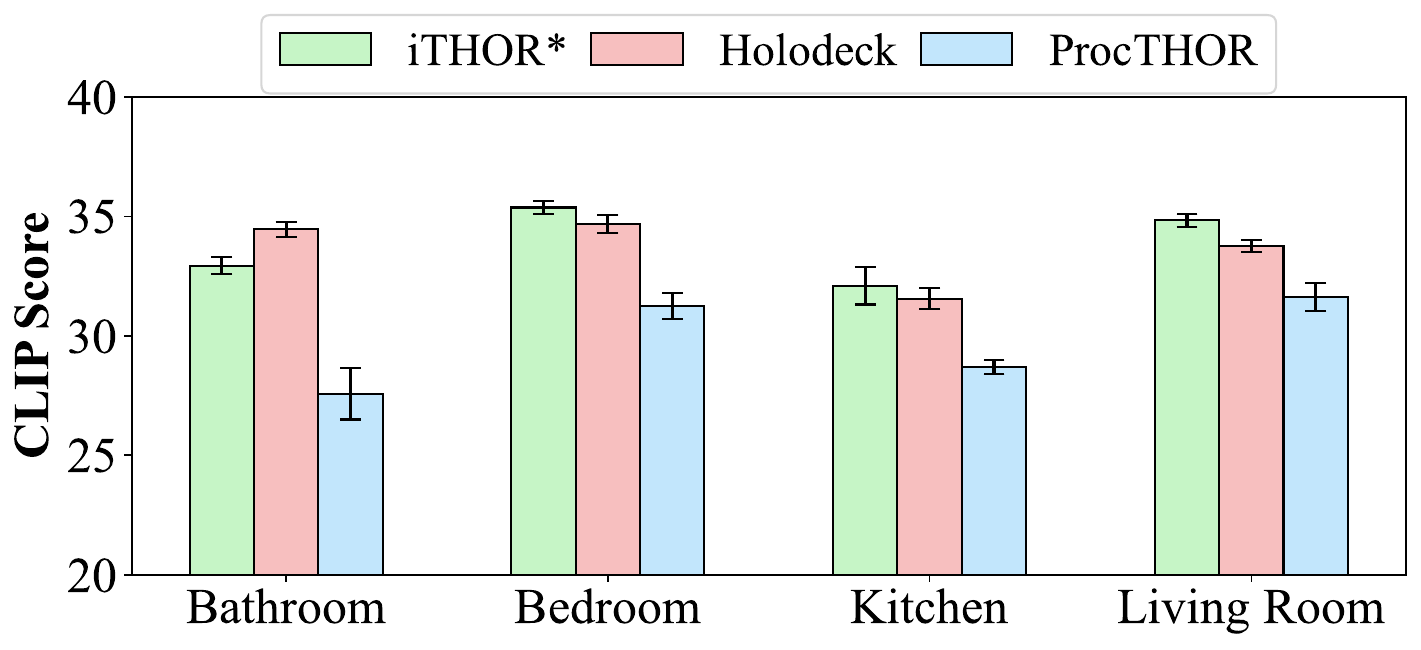}
    \caption{CLIP Score comparison over four residential scene types. * denotes iTHOR scenes are designed by human experts.}
    \label{fig: clip_score}
    \vspace{-0.5cm}
\end{figure}

\begin{figure*}[!t]
\centering
\includegraphics[width=\textwidth]{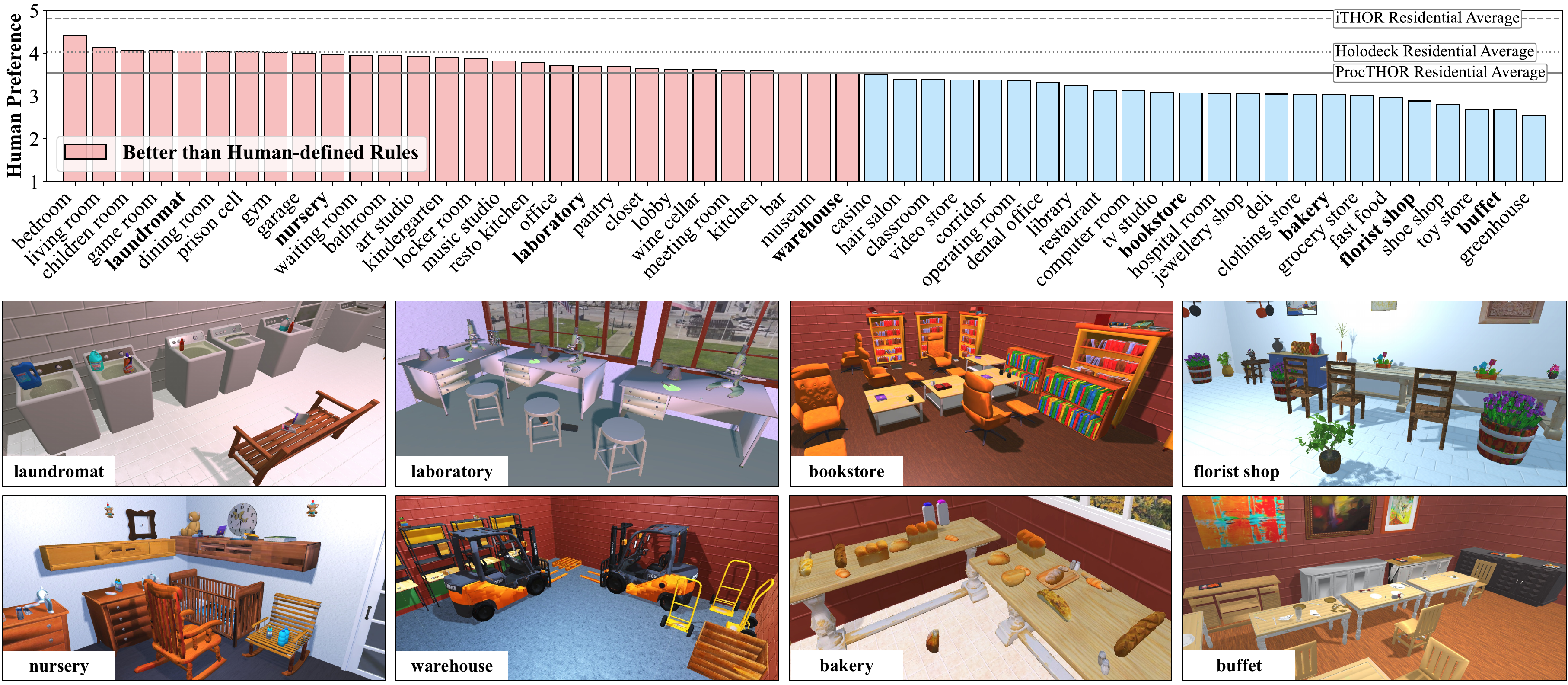}
    \caption{Human evaluation on 52 scene types from MIT Scenes \cite{quattoni2009recognizing} with qualitative examples. The three horizontal lines represent the average score of iTHOR, \holodeck, and \procthor on four types of residential scenes (\textit{bedroom}, \textit{living room}, \textit{bathroom} and \textit{kitchen}.)}
    \label{fig: mit scenes}
    \vspace{-0.4cm}
\end{figure*}
In addition to human judgments, we employ CLIP Score\footnote{Here, we use OpenCLIP \cite{ilharco_gabriel_2021_5143773} with ViT-L/14 trained on LAION-2B \cite{schuhmann2022laionb}. We use cosine similarity times 100 as the CLIP Score.} \cite{hessel2021clipscore} to quantify the visual coherence between the top-down view of the scene and its corresponding scene type embedded in a prompt template ``\textit{a top-down view of [scene type]}''. Besides, we add human-designed scenes from iTHOR \cite{ai2thor} as the upper bound for reference. Figure \ref{fig: clip_score} shows the CLIP scores of \holodeck exceed \procthor with great margins and closely approach the performance of iTHOR, demonstrating \holodeck's ability to generate visually coherent scenes faithful to the designated scene types. The CLIP Score experiment agrees with our human evaluation.

\subsection{\textbf{\holodeck} on Diverse Scenes}
To evaluate \holodeck's capability beyond residential scenes, we have humans rate its performance on 52 scene types\footnote{Limited by the \procthor framework, we filter those scenes types that require special structures such as \textit{swimming pool}, \textit{subway}, etc.} from MIT Scenes Dataset~\cite{quattoni2009recognizing}, covering five categories: Stores (\textit{deli}, \textit{bakery}), Home (\textit{bedroom}, \textit{dining room}), Public Spaces (\textit{museum}, \textit{locker room}), Leisure (\textit{gym}, \textit{casino}) and Working Space (\textit{office}, \textit{meeting room}).
\smallbreak
\noindent \textbf{Setup.} We prompt \holodeck to produce five outputs for each type using only the scene name as the input, accumulating 260 examples across the 52 scene types. Annotators are presented with a top-down view image and a 360-degree video for each scene and asked to rate them from 1 to 5 (with higher scores indicating better quality), considering asset selection, layout coherence, and overall match with the scene type. To provide context for these scores, we include residential scenes from \procthor and iTHOR in this study, with 20 scenes from each system.
\smallbreak
\noindent \textbf{\holodeck can generate satisfactory outputs for most scene types.}
Figure~\ref{fig: mit scenes} demonstrates the human preference scores for diverse scenes with qualitative examples. Compared to \procthor's performance in residential scenes, \holodeck achieves higher human preference scores over half of (28 out of 52) the diverse scenes. Given that \procthor relies on human-defined rules and residential scenes are relatively easy to build with common objects and simple layout, \holodeck's breadth of competence highlights its robustness and flexibility in generating various indoor environments. However, we notice that \holodeck struggles with scenes requiring more complex layouts such as \textit{restaurant} or unique assets unavailable in Objaverse, e.g., ``a dental x-ray machine'' for the scene \textit{dental office}.
Future work can improve the system by incorporating more assets and introducing more sophisticated layout algorithms.
\subsection{Ablation Study on Layout Design} \label{sec: layout ablation}
 

\begin{table}[!t]
    \centering
    \resizebox{8.3cm}{!}{%
    \begin{tabular}{cccccc}
    \toprule
     \textbf{Method}    & \textbf{Bathroom} & \textbf{Bedroom} & \textbf{Kitchen} & \textbf{Living Room} & \textbf{Average}\\ 
     \midrule
 
    \textsc{Absolute}   & 0.369 & 0.343 & 0.407  & 0.336  & 0.364\\
    \textsc{Random}    & 0.422 & 0.339  & 0.367  & 0.348  & 0.369\\
    \textsc{edge}      & 0.596  & 0.657  & \textbf{0.655} & 0.672 & 0.645\\
    \midrule
     \textsc{Constraint}   & \textbf{0.696} & \textbf{0.745}  & 0.654  & \textbf{0.728} & \textbf{0.706} \\
     \bottomrule
    \end{tabular} 
    }
    \caption{Mean Reciprocal Rank ($\uparrow$) of different layouts ranked by human. \textsc{Constraint}: using spatial relational constraints; \textsc{Absolute}: LLM-defined absolute positions; \textsc{Random}: randomly place the objects and \textsc{Edge}: put objects at the edge of the room.}
    \label{tab:placement_abalation}
    \vspace{-0.4cm}
\end{table}

This user study aims to validate the effectiveness of \holodeck's constraint-based layout design method.
\smallbreak
\noindent \textbf{Baselines.} We consider four layout design methods: (1) \textsc{Constraint}: the layout design method of \holodeck; (2) \textsc{Absolute}: directly obtaining the absolute coordinates and orientation of each object from LLM akin to LayoutGPT \cite{feng2023layoutgpt}; (3) \textsc{Random}: randomly place all objects in the room without collision; (4) \textsc{Edge}: placed objects along the walls.\looseness=-1
\smallbreak
\noindent \textbf{Setup.} We modify the residential scenes of \holodeck used in \ref{sec: holodeck vs procthor} by altering the layouts using the previously mentioned methods while keeping the objects in the scene identical. We present humans with four shuffled top-down images from each layout strategy and ask them to rank the four layouts considering out-of-boundary, object collision, reachable space, and layout realism.
\smallbreak
\begin{table*}[!t]
\small
\centering
\resizebox{17cm}{!}{%
\begin{tabular}{lcccccccccccc}
\toprule
\textit{}& \multicolumn{2}{c}{\textbf{Office}}   & \multicolumn{2}{c}{\textbf{Daycare}}  & \multicolumn{2}{c}{\textbf{Music Room}}& \multicolumn{2}{c}{\textbf{Gym}}    & \multicolumn{2}{c}{\textbf{Arcade}} & \multicolumn{2}{c}{\textbf{Average}}   \\ 
\cmidrule{2-13}
\multicolumn{1}{l}{\textbf{Method}} & \multicolumn{1}{c}{Success} & \multicolumn{1}{c}{SPL} & \multicolumn{1}{c}{Success} & \multicolumn{1}{c}{SPL} & \multicolumn{1}{c}{Success} & \multicolumn{1}{c}{SPL} & \multicolumn{1}{c}{Success} & \multicolumn{1}{c}{SPL} & \multicolumn{1}{c}{Success} & \multicolumn{1}{c}{SPL} & \multicolumn{1}{c}{Success} & \multicolumn{1}{c}{SPL}\\ 
\midrule
Random & 3.90 & 0.039 & 4.05 & 0.041 & 5.20 & 0.052 & 2.84 & 0.029 & 2.54 & 0.025 & 3.71 & 0.037 \\
\procthor~\cite{procthor} & 8.77 & 0.031 & 2.87 & 0.011 & 6.17 & 0.027 & 0.68 & 0.002 & 2.06 & 0.005 & 4.11 & 0.015  \\
${+}$\textsc{Objaverse} (ours) & 18.42 & 0.068 & 8.99 & 0.061 & 25.69 & 0.157 & \textbf{18.79} & 0.101 & \textbf{13.21} & \textbf{0.076} & 17.02 & 0.093   \\
${+}$\holodeck (ours) & \textbf{25.05} & \textbf{0.127} & \textbf{15.61} & \textbf{0.127} & \textbf{31.08} & \textbf{0.202} & 18.40 & \textbf{0.110} & 11.84 & 0.069 & \textbf{20.40} & \textbf{0.127}
\\ 
\bottomrule
\end{tabular}
}
\caption{Zero-shot ObjectNav on \noveltythor. \procthor is the model pretrained on \textsc{ProcTHOR-10K} \cite{procthor}. ${+}$\textsc{Objaverse} and ${+}$\textsc{Holodeck} stand for models finetuned on the corresponding scenes. We report Success (\%) and Success weighted by Path Length (SPL).}
\label{tab: embodied_results}
\vspace{-0.5cm}
\end{table*}
\begin{figure}[!t]
\centering
\includegraphics[width=8.3cm]{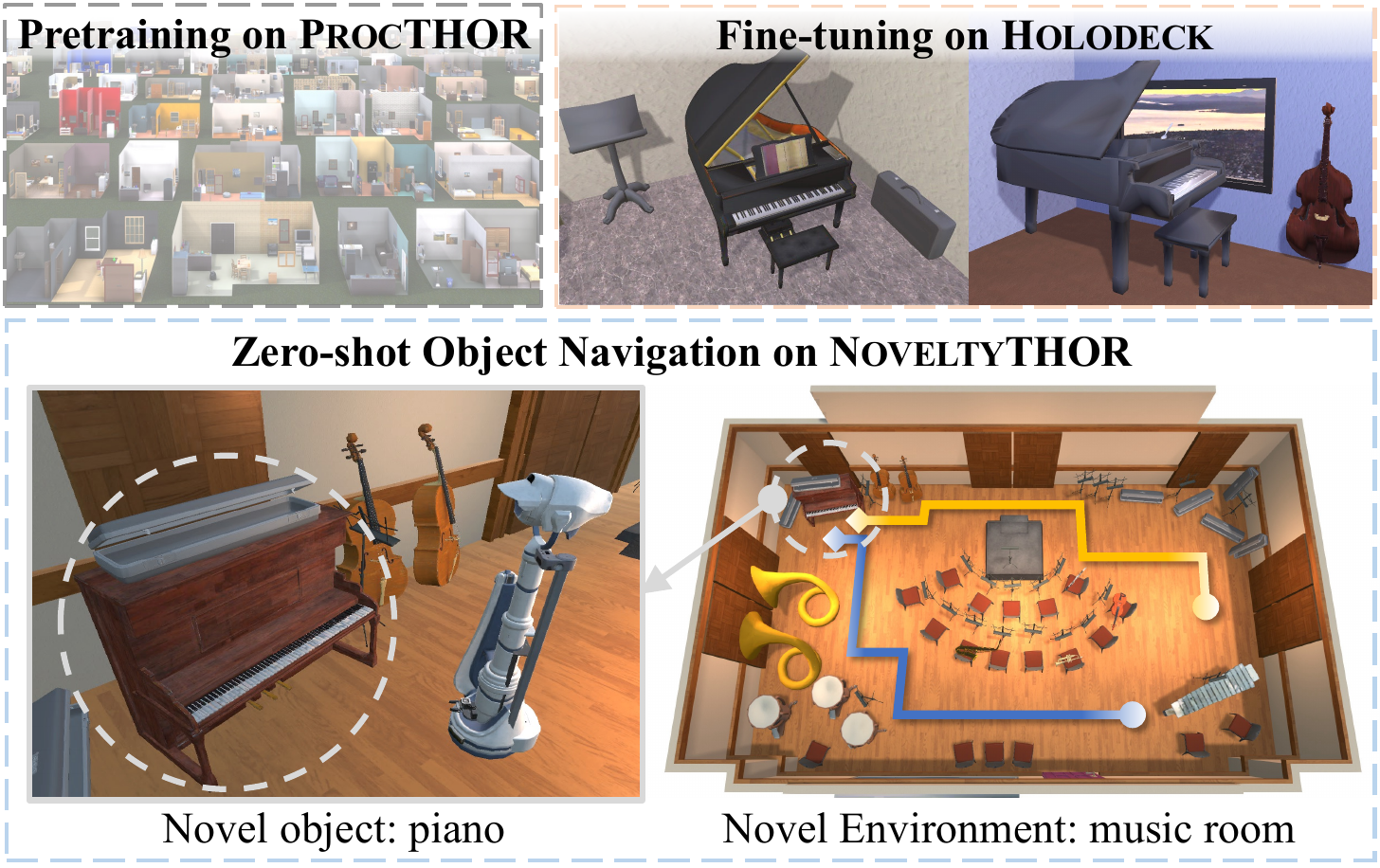}
    \caption{Zero-shot object navigation in novel scenes. Given a novel scene type, e.g., \textit{Music Room}, \holodeck can synthesize new scenes for fine-tuning to improve the performance of pretrained agents in expert-designed environments.}
    \label{fig: embodied demo}
    \vspace{-0.4cm}
\end{figure}
\noindent \textbf{Constraint-based layout is more reliable.}
Table \ref{tab:placement_abalation} reports the Mean Reciprocal Rank of different layout design methods. \holodeck's constraint-based approach outperforms the other methods significantly on \textit{bathroom}, \textit{bedroom} and \textit{living room}. \textsc{Constraint} and \textsc{Edge} perform similarly on \textit{kitchen}, where it is common to align most objects against walls. The \textsc{Absolute} method performs no better than \textsc{Random} due to its tendency to create scenes with collision and boundary errors (see examples in the supplement), typically rated poorly by humans. These results endorse spatial relational constraints as a viable strategy for generating scenes that adhere to commonsense logic.

\section{Object Navigation in Novel Environments}
As illustrated in Figure \ref{fig: embodied demo}, one application of \holodeck is synthesizing training environments to better match a novel testing distribution. To study this application, we consider ObjectNav~\cite{batra2020objectnav}, a common task in which a robot must navigate toward a specific object category. As existing benchmarks~\cite{deitke2020robothor, procthor,ramakrishnan2021hm3d} for ObjectNav consider only household environments and support a very limited collection of object types (16 object types in total combining the above benchmarks), we introduce \noveltythor, an artist-designed benchmark to evaluate embodied agents in diverse environments. Subsequently, we use the ObjectNav model pre-trained on \textsc{ProcTHOR-10K}~\cite{ai2thor} and finetune it on 100 scenes generated by \holodeck. These scenes are created by prompting \holodeck with the novel scene type as input. The model is then evaluated on \noveltythor.

\smallbreak
\noindent \textbf{\noveltythor.} We have two professional digital artists manually create 10 novel testing environments with two examples for each of the five categories: \textit{Office}, \textit{Daycare}, \textit{Music Room}, \textit{Gym}, and \textit{Arcade}. 
Each scene contains novel object types not included in the existing ObjectNav tasks, e.g., ``piano'' in \textit{Music Room}, ``treadmill'' in \textit{Gym}, etc. Across \noveltythor, there are 92 unique object types.
\smallbreak

\noindent \textbf{Baselines.} For all methods except the one of random action, we use the same pre-trained ObjectNav model from \textsc{ProcTHOR-10K}~\cite{ai2thor}, which has been trained for ${\approx}$400M steps to navigate to 16 object categories. 
To adapt the agent to novel scenes without human-construct training data, we consider two methods: (1) $+$\holodeck: we prompt\footnote{Here, we prompt with the scene name and its paraphrases to get more diverse outputs, e.g., we use ``game room'', ``amusement center'' for \textit{Arcade}.} \holodeck to generate 100 scenes for each scene type automatically; (2) $+$\textsc{Objaverse}: a strong baseline by enhancing \procthor with \holodeck's scene-type-specific object selection, specifically, those scenes are populated with similar Objaverse assets chosen by \holodeck.





\smallbreak
\noindent \textbf{Model.} Our ObjectNav models use the CLIP-based architectures of \cite{khandelwal2022:embodied-clip}, which contains a CNN visual encoder and a GRU to capture temporal information. We train each model with 100 scenes for 50M steps, which takes approximately one day on 8 Quadro RTX 8000 GPUs. We select the checkpoint of each model based on the best validation performance on its own validation scenes.
\smallbreak
\noindent \textbf{Results.} Table \ref{tab: embodied_results} shows zero-shot performance on \noveltythor. \holodeck achieves the best performance on average and surpasses baselines with considerable margins on \textit{Office}, \textit{Daycare}, and \textit{Music Room}. On \emph{Gym} and \emph{Arcade}, $+$\textsc{Holodeck} and $+$\textsc{Objaverse} perform similarly. Given that the main difference between $+$\textsc{Holodeck} and $+$\textsc{Objaverse} scenes is in the object placements, the observed difference suggests that \holodeck is more adept at creating layouts that resemble those designed by humans. For example, We can observe in Figure~\ref{fig: embodied demo} that the music room in \noveltythor contains a piano, violin cases, and cellos that are in close proximity to each other. The music room generated by \holodeck also shows a similar arrangement of these objects, highlighting the ``commonsense'' understanding of our method.
\procthor struggles in \noveltythor, often indistinguishably from random, because of poor object coverage during training.



\section{Conclusion and Limitation}
We propose \holodeck, a system guided by large language models to generate diverse and interactive Embodied AI environments with text descriptions. We assess the quality of \holodeck with large-scale human evaluation and validate its utility in Embodied AI through object navigation in novel scenes. We plan to add more 3D assets to \holodeck and explore its broader applications in Embodied AI in the future.

{
    \small
    \bibliographystyle{ieeenat_fullname}
    \bibliography{main}
}

\clearpage
\appendix
\section{Details of \holodeck} \label{appendix: holodeck details}

\subsection{Efficiency and Cost}
To create an interactive house of $k$ rooms, \holodeck uses $3 + 3 \times k$ API calls. More specifically, utilizing OpenAI's \href{https://platform.openai.com/docs/models/gpt-4-and-gpt-4-turbo}{gpt-4-1106-preview} model incurs an approximate cost of \$ 0.2 per room. With our current implementation, \holodeck can generate a single room in about 3 minutes. This includes the time for API calls and layout optimization using a MacBook equipped with an M1 chip.

\subsection{Floor \& Wall Modules}
In the LLM outputs in the Floor Module, the following details are provided for each room:
\begin{itemize}
    \item \textbf{room type}: the room's name,  e.g., kitchen, bedroom.
    \item \textbf{floor material}: a description of the floor's appearance.
    \item \textbf{wall material}: a description of the wall's appearance.
    \item \textbf{vertices}: four tuples $\{(x_i, y_i), i \in [1, 2, 3, 4]\}$, representing the coordinates of the room's corners.
\end{itemize}
\noindent \textbf{Material Selection.}  We have an image representation for each of 236 materials, consistent with the material setup in \procthor~\cite{procthor}\footnote{Procthor splits the set of materials into wall and floor materials. For \holodeck, we merge them in one pool for retrieval.}. Using CLIP\footnote{We employ OpenCLIP \cite{ilharco_gabriel_2021_5143773} with ViT-L/14, trained on the LAION-2B dataset \cite{schuhmann2022laionb}, for all CLIP-related components in this paper.} \cite{radford2021learning}, we calculate the similarity between the material descriptions provided by the Large Language Model (LLM) and these images. The material with the highest similarity score is selected. Additionally, we utilize the 148 colors from Matplotlib~\cite{Hunter:2007} to refine the material selection by choosing the color closest to the description with CLIP.
\smallbreak
\noindent \textbf{Wall height.} We have the LLM suggest a suitable wall height based on the user's input. For example, it may recommend a high ceiling for commercial spaces like museums.

\subsection{Doorway \& Window Modules}
\begin{figure}[!t]
\centering
\includegraphics[width=8.3cm]{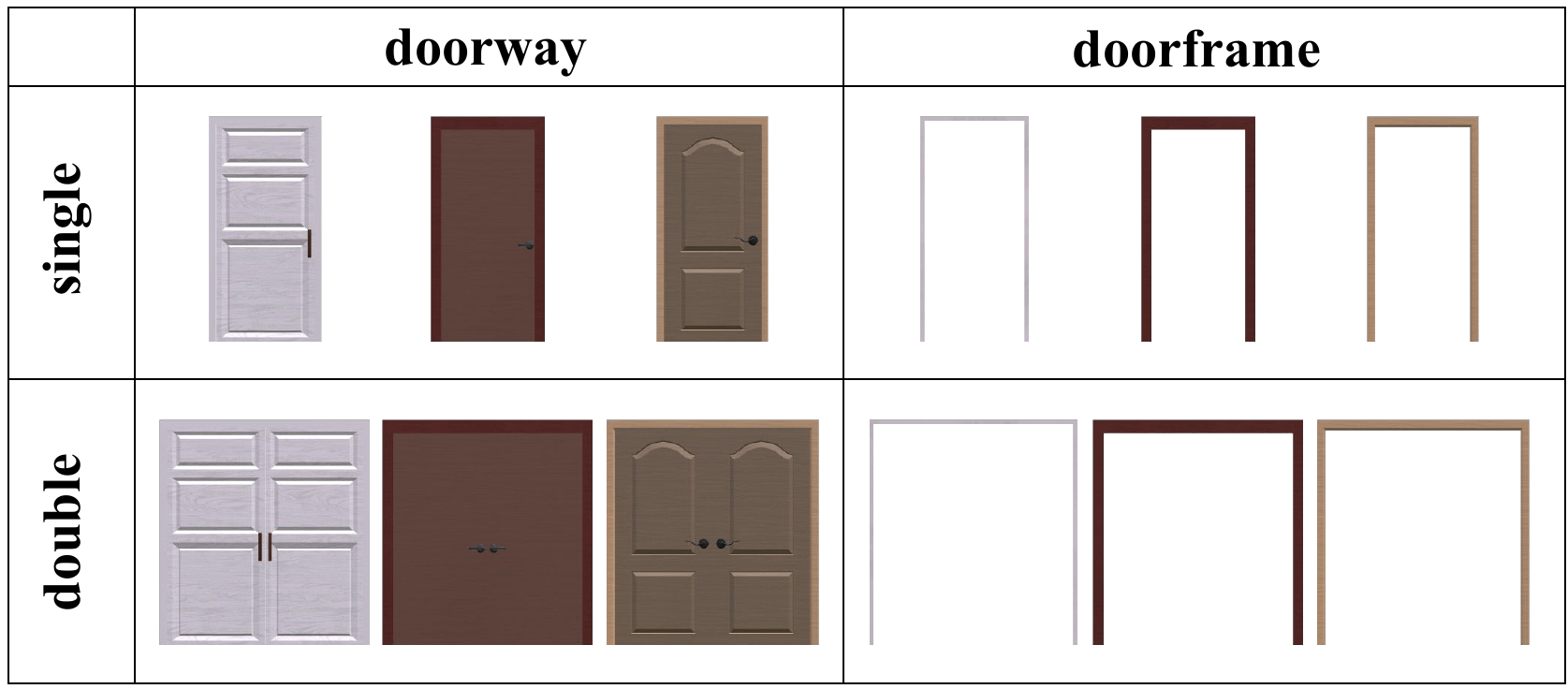}
    \caption{Examples of different doors in \holodeck.}
    \label{fig: door_assets}
    \vspace{-0.4cm}
\end{figure}

\begin{figure}[!t]
\centering
\includegraphics[width=8.3cm]{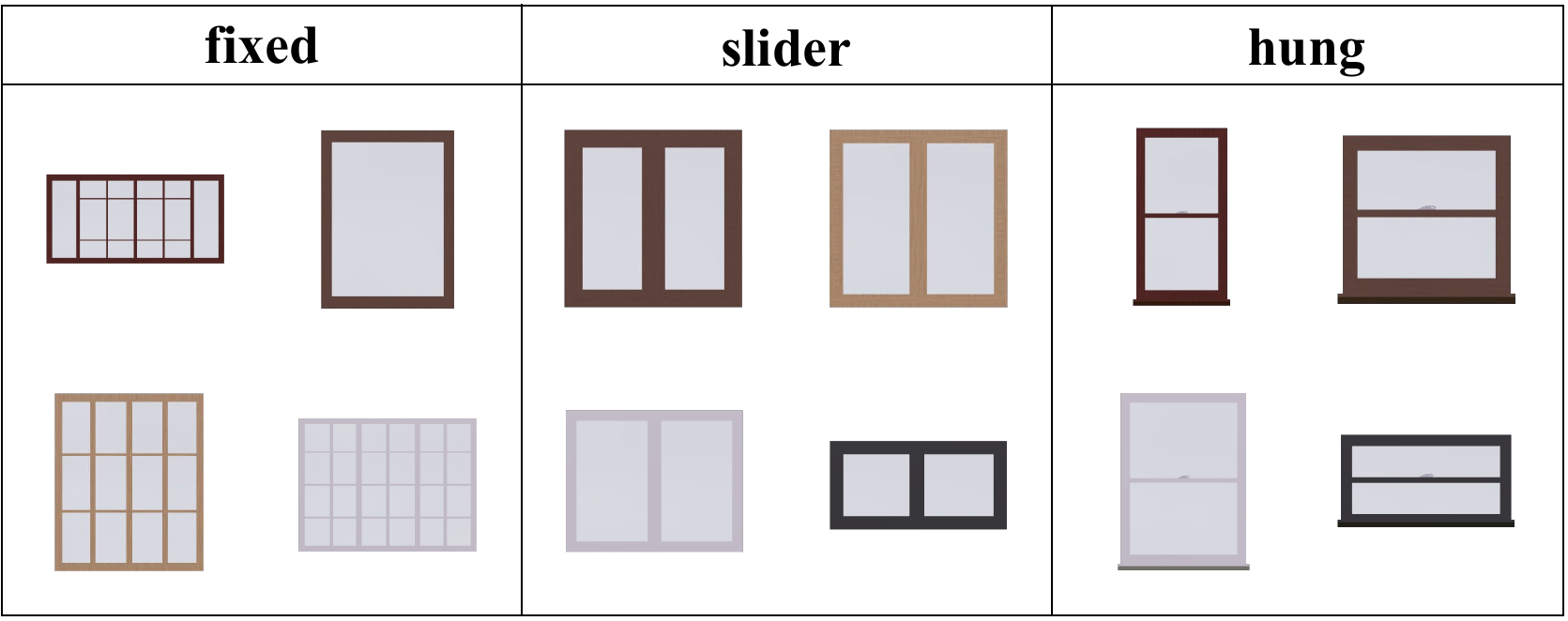}
    \caption{Examples of different windows in \holodeck.}
    \label{fig: window_assets}
    \vspace{-0.4cm}
\end{figure}

In \holodeck, we take advantage of the diverse collection of doors and windows introduced in \procthor~\cite{procthor}, featuring a diverse collection of 40 doors (refer to examples in Figure \ref{fig: door_assets}) and 21 windows (see Figure \ref{fig: window_assets}). The LLM provides essential information to aid in the selection of doors:
\begin{itemize}
    \item \textbf{room 1 \& room 2}: the two rooms connected by the door, for example, bedroom and kitchen.
    \item \textbf{connection type}: one of the three connection types: \textit{doorframe} (frame without a door), \textit{doorway} (frame with a door), and \textit{open} (no wall separating the rooms).
    \item \textbf{size}: the size of the door: \textit{single} (one meter in width) or \textit{double} (two meters in width).
    \item \textbf{door style}: a description of the door's appearance.
\end{itemize}

\noindent We have an image for each door, and we utilize CLIP to select the door that most closely matches the description.
\smallbreak
\noindent We have the LLM provide the following data about windows:
\begin{itemize}
    \item \textbf{room type}: the room where the window will be installed.
    \item \textbf{direction}: the wall's direction (\textit{south}, \textit{north}, \textit{east}, or \textit{west} ) where the window will be placed.
    \item \textbf{type}: one of the three window types: \textit{fixed}, \textit{slider} or \textit{hung}.
    \item \textbf{size}: the width and height of the window.
    \item \textbf{quantity}: the number of windows installed on each wall.
    \item \textbf{height}: the distance from the floor to the window's base.
\end{itemize}

\begin{figure*}[!t]
\centering
\includegraphics[width=\textwidth]{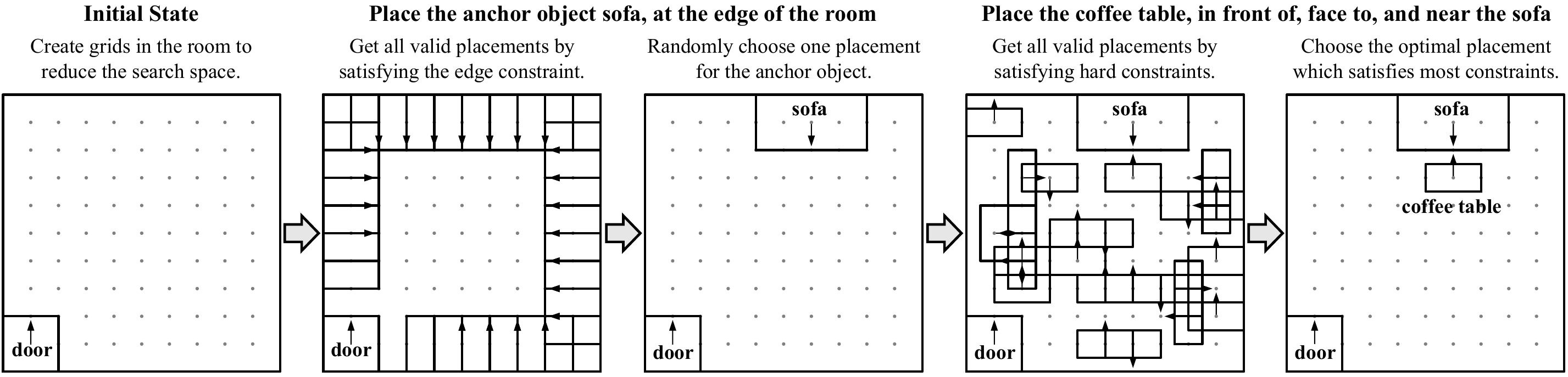}
    \caption{Example of using the DFS-based Constraint Satisfaction algorithm to place the objects.}
    \label{fig: dfs_demo}
    \vspace{-0.4cm}
\end{figure*}

\subsection{Object Selection Module} \label{sec: object selection module}

In Objaverse, each 3D asset $o \in \mathcal{O}$ is associated with the following metadata - a textual description of the asset $t$, the 3D bounding box size of the asset $\left(w, d, h\right)$, and a set of 2D images $I$ captured from three different angles ($0\degree$, $45\degree$, and $-45\degree$). 
For each object proposed by LLM $o'$, we have the LLM output a detailed description of the object ($t'$) and its 3D bounding box size $(w', d', h')$ for retrieval purposes.
To evaluate the similarity between a candidate 3D asset in the repository $o = \left(t,  \left(w, d, h\right), I\right)$ and the object proposed by the LLM $o' \left( t', \left(w', d', h'\right)\right)$, we use three metrics:

\begin{itemize}
    \item \textbf{Visual Similarity} ($\mathcal{V}$) measures the CLIP similarity between the 2D renderings of the candidate asset and the textual description of the LLM-proposed object: $\mathcal{V}(o, o') = \max_{i \in I}\text{CLIP}(i, t')$.
    \item \textbf{Textual Similarity} ($\mathcal{T}$) measures the similarity between the textual description of the candidate 3D asset and the textural description of the LLM-proposed object. This metric is crucial in improving the accuracy of the retrieval process since it ensures that we retrieve the asset within the correct category. We use the sentence transformer (SBERT) \cite{reimers-2019-sentence-bert} with \href{https://huggingface.co/sentence-transformers/all-mpnet-base-v2}{all-mpnet-base-v2} checkpoint to calculate the scores: $\mathcal{T} = \text{SBERT}(t, t')$.
    \item \textbf{Size Discrepancy} ($\mathcal{S}$) measures the discrepancy in the size of the 3D bounding box size of the candidate asset and the LLM-proposed object. There are similar objects with different sizes in the asset repository, and the size of objects is an important factor in designing scenes, e.g., we need a larger sofa for a large living room. The size matching score is computed as: $\mathcal{S}(o, o') = \left(|w-w'| + |h - h'| + |d - d'|\right)/3$. Two objects of similar size will have a smaller value of $\mathcal{S}$.
\end{itemize}

The overall matching score $\mathcal{M} \left(o, o'\right)$ is a weighted sum of the above metrics:
\begin{equation}
    \mathcal{M} \left(o, o'\right) = \alpha \cdot \mathcal{V}\left(o, o'\right) + \beta \cdot \mathcal{T} (o, o') - \gamma \cdot \mathcal{S} (o, o')
\end{equation}
with weights $\alpha = 100$, $\beta = 1$, and $\gamma = 10$. The asset with the highest matching score is selected.

\subsection{Layout Design Module} \label{sec: object placement module}
In this module, we position the set of objects $O$ chosen in Sec \ref{sec: object selection module}, applying spatial relational constraints provided by the LLM. We define various constraints for floor objects:
\begin{itemize}
    \item \textbf{Global constraint}: edge; middle.
    \item \textbf{Distance constraint}: near (object); far (object).
    \item \textbf{Position constraint}: in front of (object); side of (object).
    \item \textbf{Alignment constraint}: center align with (object).
    \item \textbf{Direction constraint}: face to (object).
\end{itemize}

The LLM can combine these constraints to form a constraint list $C_o$ for each object $o \in O$. For instance, as shown in Figure \ref{fig: dfs_demo}, the constraints for a ``coffee table'' are [\text{middle}, \text{in front of (sofa)}, \text{face to (sofa)}, \text{near (sofa)}].

For floor object placement, we employ two solvers: Depth-First-Search (DFS) Solver and Mixed Integer Linear Programming (MILP) \cite{benichou1971experiments} Solver.
\smallbreak
\noindent \textbf{Depth-First-Search Solver.} In the DFS solver, each object is defined by five variables $\left(x, y, w, d, \text{rotation}\right)$. $\left(x, y\right)$ is the 2D coordinates of the object's center, $w$ and $d$ are the width and depth of the 2D bounding box of the object, and rotation can be one of 0°, 90°, 180°, and 270°.
The constraints listed above are treated softly, allowing certain violations when finding a layout.
Beyond these soft constraints, we implement hard constraints essential for object placement: these constraints prevent object collisions and ensure that objects remain within the designated room boundaries. Violation of these hard constraints results in the object not being placed. Figure \ref{fig: dfs_demo} demonstrates that our DFS solver initiates grids to establish a finite search space. It first explores different placements for the anchor object selected by the LLM. Subsequent steps involve optimizing the placement for the remaining objects, adhering to the hard constraints, and satisfying as many soft constraints as possible.\footnote{The evaluation of an object's placement is based on the number of constraints satisfied. Placements that satisfy a greater number of constraints receive higher weights. However, any placement that violates hard constraints is rejected.} The algorithm can yield multiple solutions, with the final selection meeting the most constraints.
\smallbreak
\noindent \textbf{Mixed Integer Linear Programming (MILP) Solver} is particularly effective for structured layout design. It optimizes a linear objective function subject to linear constraints with some non-discrete variables. This approach is well-suited for our layout optimization problem in \holodeck.

In our MILP formulation, each object's position is determined by four variables: ($x$, $y$, $\text{rotate}_{90}$, $\text{rotate}_{180}$). The variables $\text{rotate}_{90}$ and $\text{rotate}_{180}$ are boolean, indicating rotations of 90 and 180 degrees, respectively. For example, if $\text{rotate}_{90}$ and $\text{rotate}_{180}$ are both true, it signifies a 270-degree rotation of the object.
We translate all previously mentioned constraints into linear ones for the MILP problem. For instance, to align Object A with Object B at the center, a constraint in the form of \(A_x = B_x\) or \(A_y = B_y\) is implemented, where \(A_x, A_y\) and \(B_x, B_y\) represent the centers of Objects A and B, respectively. Note that the constraint is non-linear due to the OR operator. To model this linearly in MILP, we can introduce binary auxiliary variables and additional constraints to capture the logic of the OR condition.
For solving the MILP, we utilize GUROBI \cite{gurobi}, a state-of-the-art solver known for its efficiency and robustness. 

In MILP solver, all constraints specified in the previous section are applied as hard constraints except that the Distance constraints (\textit{near} and \textit{far}) are uniquely modeled as part of the objective. For a visual comparison of these solvers' outcomes in \holodeck, refer to Figure~\ref{fig: layout}.
\smallbreak
\noindent \textbf{Wall \& Small Objects.} The placement of wall objects is determined by two specific attributes:
\begin{itemize}
\item \textbf{Above (Floor Object)}: This denotes the floor object directly underneath the wall object.
\item \textbf{Height}: Specifies the exact distance from the floor to the base of the wall object, measured in centimeters.
\end{itemize}

To place small surface objects on top of larger objects, we first have LLM propose the placements and utilize\texttt{RandomSpawn}\footnote{\href{https://ai2thor.allenai.org/ithor/documentation/objects/domain-randomization/\#initial-random-spawn}{AI2-THOR RandomSpawn Documentation}} function in AI2-THOR. This method allows for randomized and efficient positioning of small objects on larger surfaces.

\begin{figure}[!t]
\centering
\includegraphics[width=8.3cm]{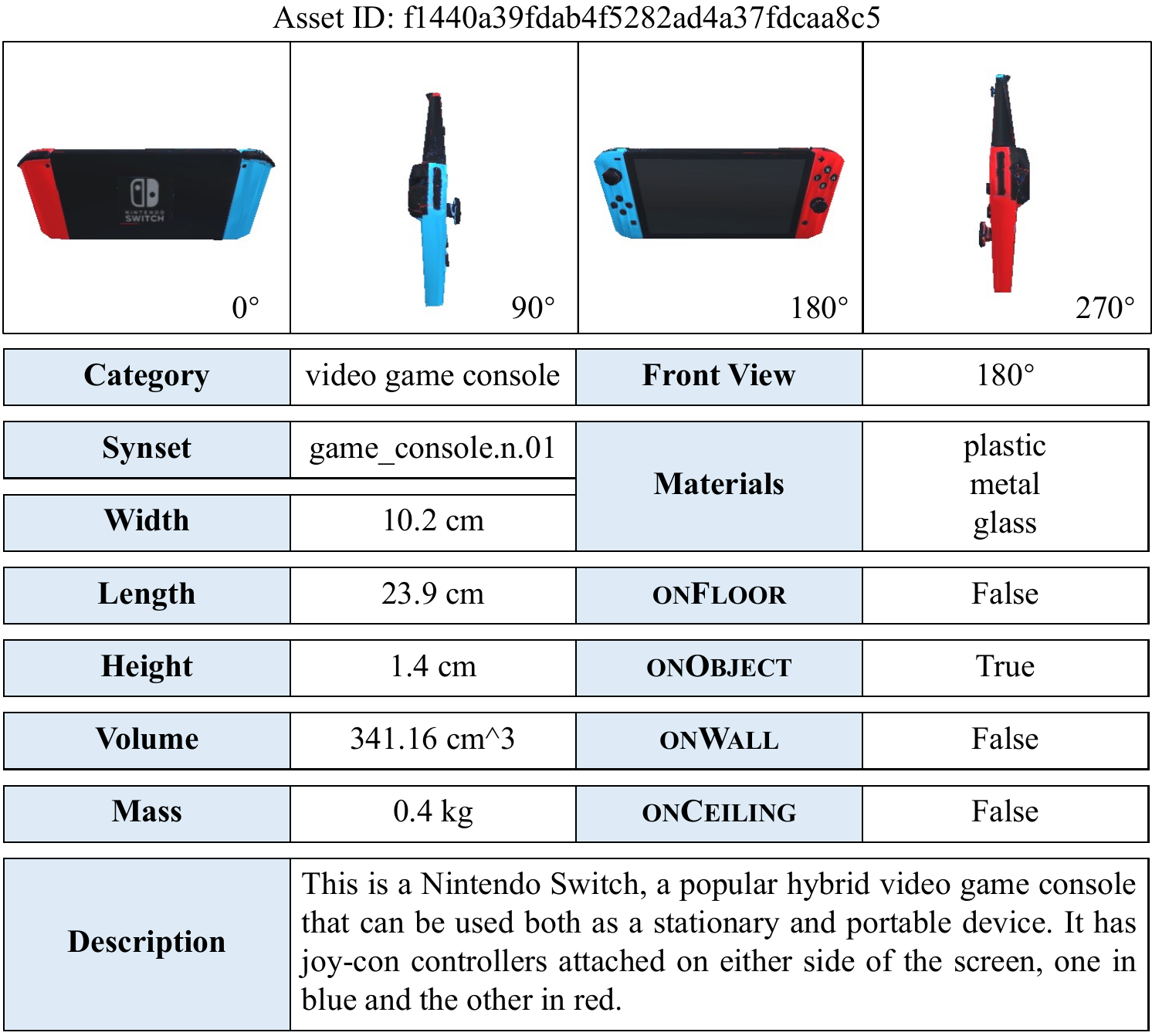}
    \caption{Example of an asset's attributes annotated by GPT-4-V.}
    \label{fig: asset_annotation}
    \vspace{-0.4cm}
\end{figure}

\subsection{GPT-4-V for 3D Asset Annotation}
We annotate the 3D assets used in \holodeck with OpenAI's \href{https://openai.com/research/gpt-4v-system-card}{GPT-4-V API} 
to enhance the accuracy of object retrieval and placement. As illustrated in Figure \ref{fig: asset_annotation}, GPT-4-V takes a set of four images as inputs, each showing an object from orthogonal rotations (0°, 90°, 180°, and 270°) and outputs the following attributes for the 3D object:
\begin{itemize}
\item \textbf{Category}: a specific classification of the object, such as ``chair'', ``table'', ``building'', etc.
\item \textbf{Synset}: the nearest WordNet \cite{miller1995wordnet} synset will be used as the object type in object navigation tasks.
\item \textbf{Width, Length, Height}: physical dimensions in centimeters, defining the object's bounding box sizes.
\item \textbf{Volume}: approximate volume in cubic centimeters (cm$^3$).
\item \textbf{Mass}: estimated object mass in kilograms (kg).
\item \textbf{Front View}: an integer denoting the view representing the front of the object, often the most symmetrical view.
\item \textbf{Description}: a detailed textual description of the object.
\item \textbf{Materials}: a list of materials constituting the object.
\item \textbf{Placement Attributes:} Boolean values (\textsc{onCeiling}, \textsc{onWall}, \textsc{onFloor}, \textsc{onObject}) indicating typical placement locations. For example, ``True'' for a ceiling fan's placement on the ceiling.
\end{itemize}

\subsection{Importing Objaverse Assets into AI2-THOR} \label{appendix: objaverse assets}
The transformation of Objaverse assets into interactive objects in AI2-THOR involves a complex, multi-step pipeline.

Initially, the process starts with downloading and converting various 3D models into a mesh format optimized for runtime loading.
We then generate visibility points on the mesh surface, enabling AI2-THOR to determine object visibility.
This is followed by 3D decomposition, where the mesh is split into simpler convex meshes to facilitate rapid and realistic collision detection.
The final step involves compressing textures (i.e., albedo, normal, and emission) and the model format to streamline performance.

Handling many assets in numerous scenes is challenging, mainly due to the large mesh counts of Objaverse assets and the traditional compile-time asset packaging approach of game engines like Unity. To address this, we implement caching layers for objects, reducing the loading time for repeated use in different scenes. Additionally, we develop a system to unload objects from memory, allowing efficient management of thousands of 3D objects at runtime.

\begin{figure}[!t]
\centering
\includegraphics[width=8.3cm]{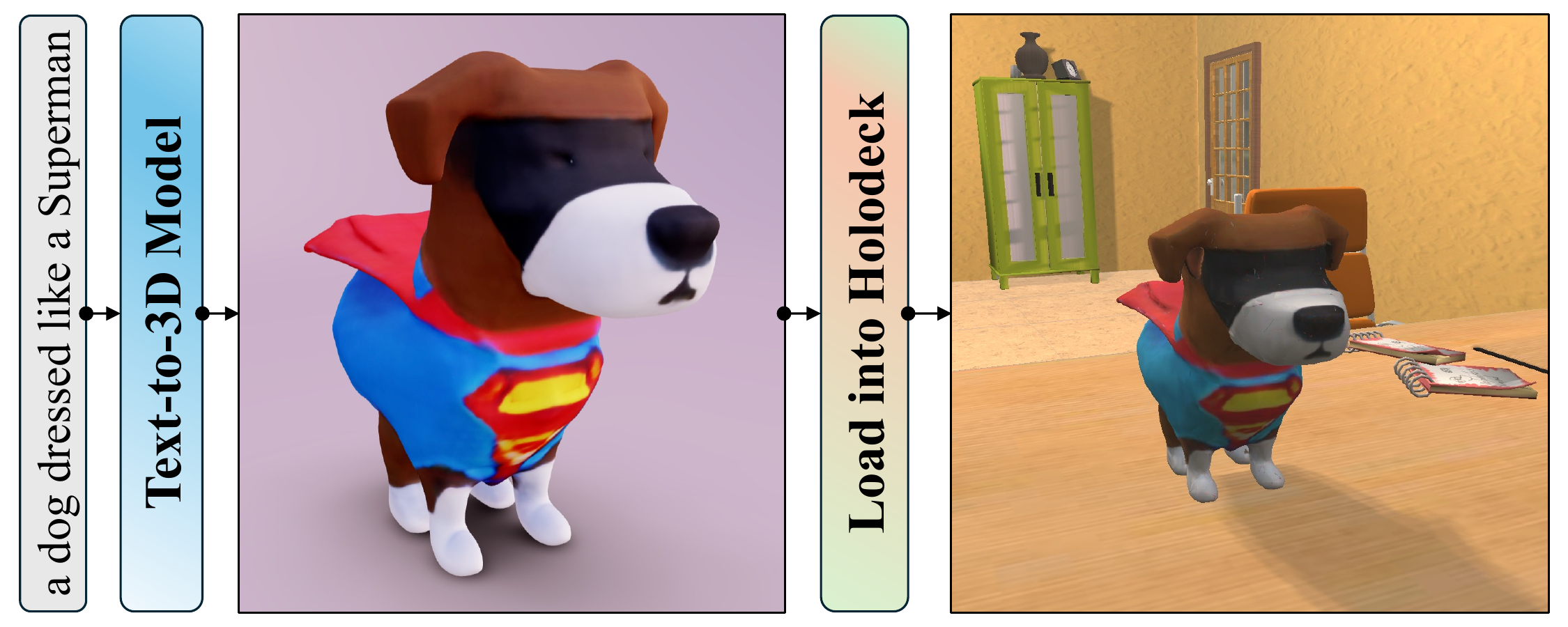}
    \caption{\holodeck can import any 3D objects, including text-to-3D models generated (e.g., the object in this figure is generated by LumaAI \cite{luma2023}) to enhance object diversity.}
    \label{fig: add new asset}
    \vspace{-0.4cm}
\end{figure}

Besides the objects from Objaverse, our automated pipeline can process any 3D objects, including those generated by text-to-3D models, as shown in Figure \ref{fig: add new asset}.

\subsection{Rendering Options}
\begin{figure}[!t]
\centering
\includegraphics[width=8.3cm]{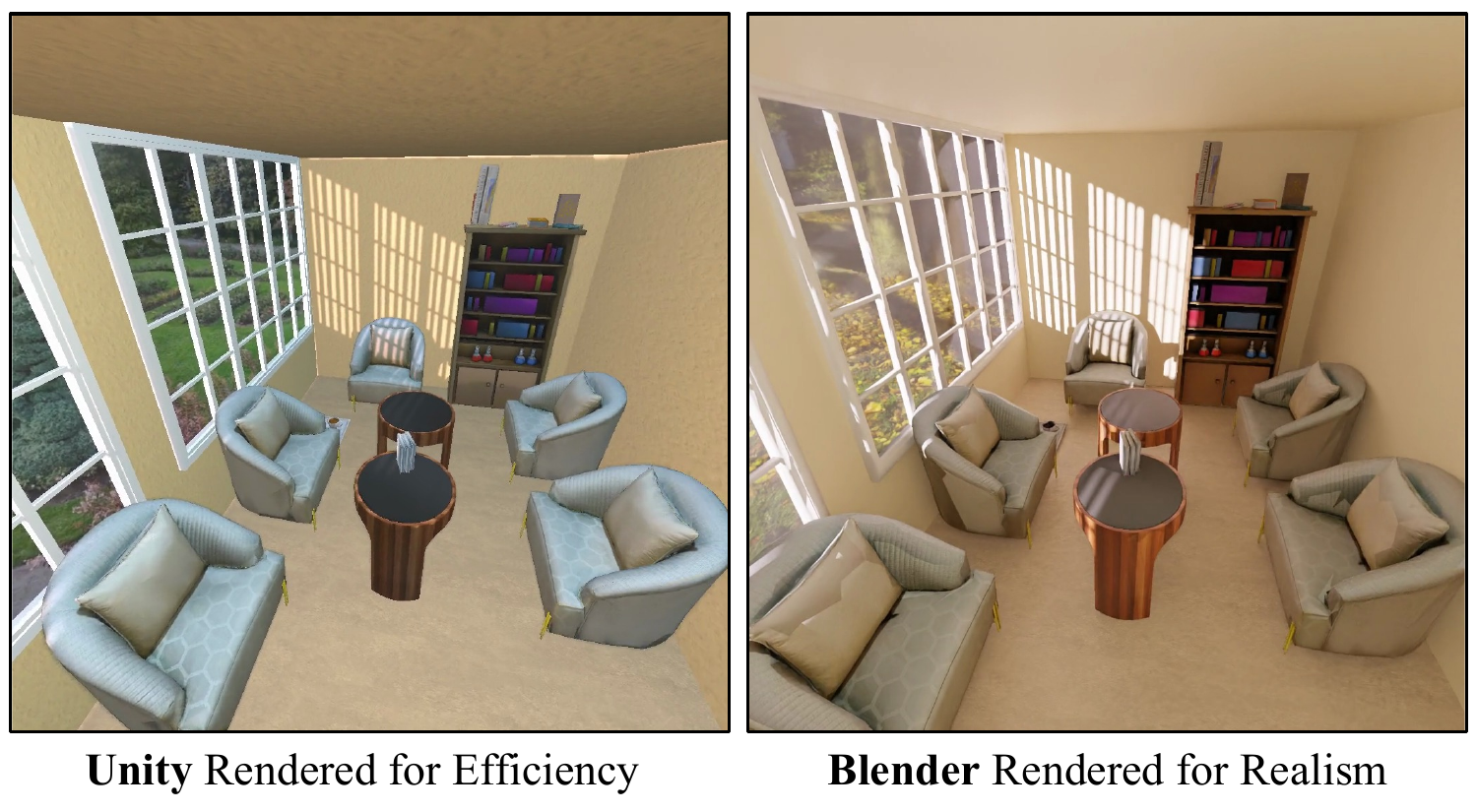}
    \caption{\holodeck renders scenes with Unity by default for efficiency to facilitate Embodied AI applications. Blender can also be used to render \holodeck scenes to improve realism.}
    \label{fig: unity_vs_blender}
\end{figure}

As shown in Figure \ref{fig: unity_vs_blender}, \holodeck scenes are rendered by Unity as default to train the embodied agents more efficiently. Users can also render \holodeck scenes in Blender to achieve better realism.

\subsection{Prompt} \label{appendix: prompt}
The complete prompt templates of \holodeck's modules are provided in Figure \ref{fig:prompt-1} and \ref{fig:prompt-2}. The prompt for annotating 3D assets using GPT-4-V is shown in Figure \ref{fig:prompt-3}.

\section{Qualitative Examples}
\begin{figure*}[!b]
\centering
\includegraphics[width=\textwidth]{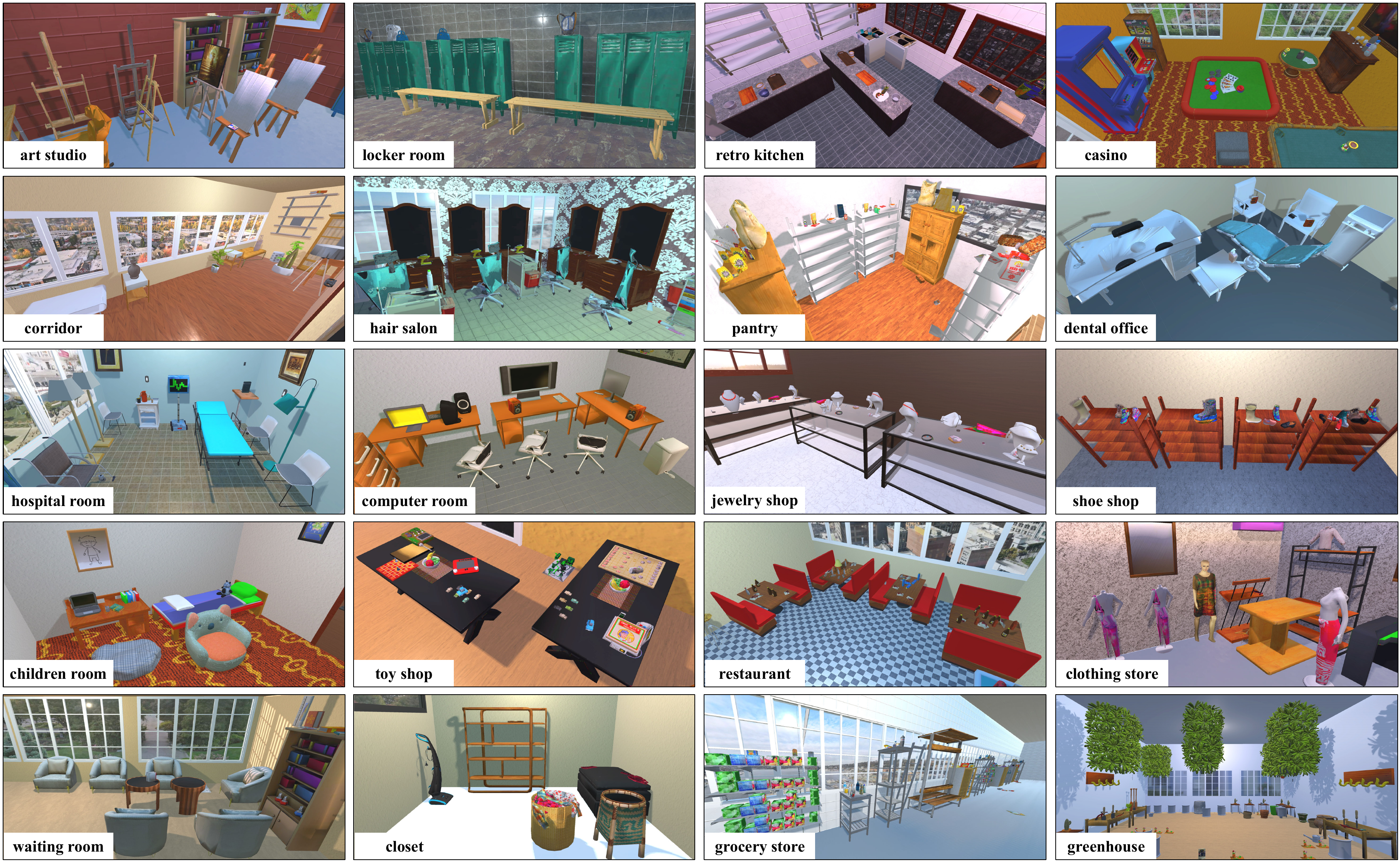}
    \caption{Additional examples of different scene types from MIT Scenes \cite{quattoni2009recognizing}.}
    \label{fig: mit_scenes_add}
\end{figure*}

\begin{figure*}[!t]
\centering
\includegraphics[width=17cm]{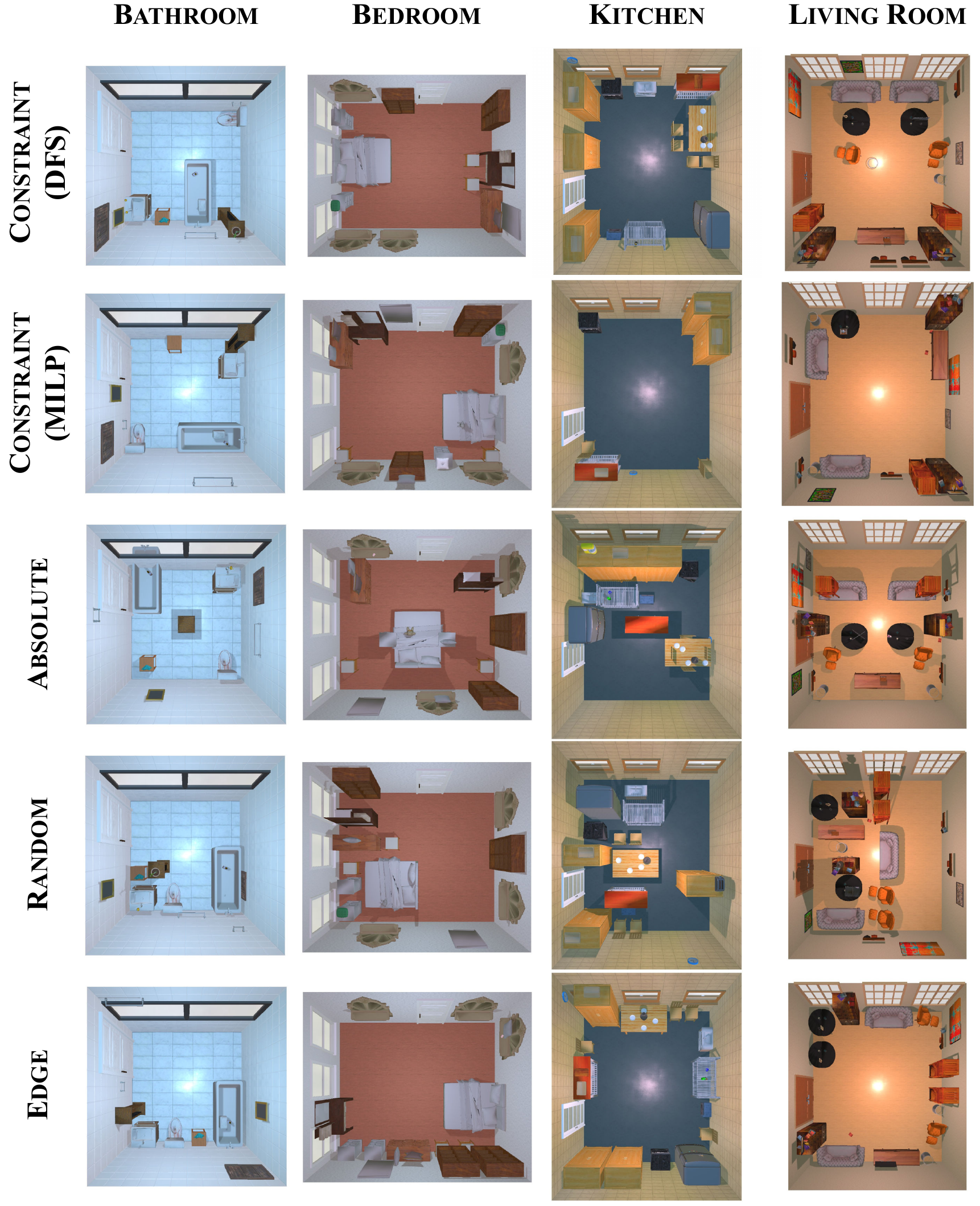}
    \caption{Qualitative comparison of the different layout methods.}
    \label{fig: layout}
\end{figure*}

\begin{figure*}[!t]
\centering
\includegraphics[width=17cm]{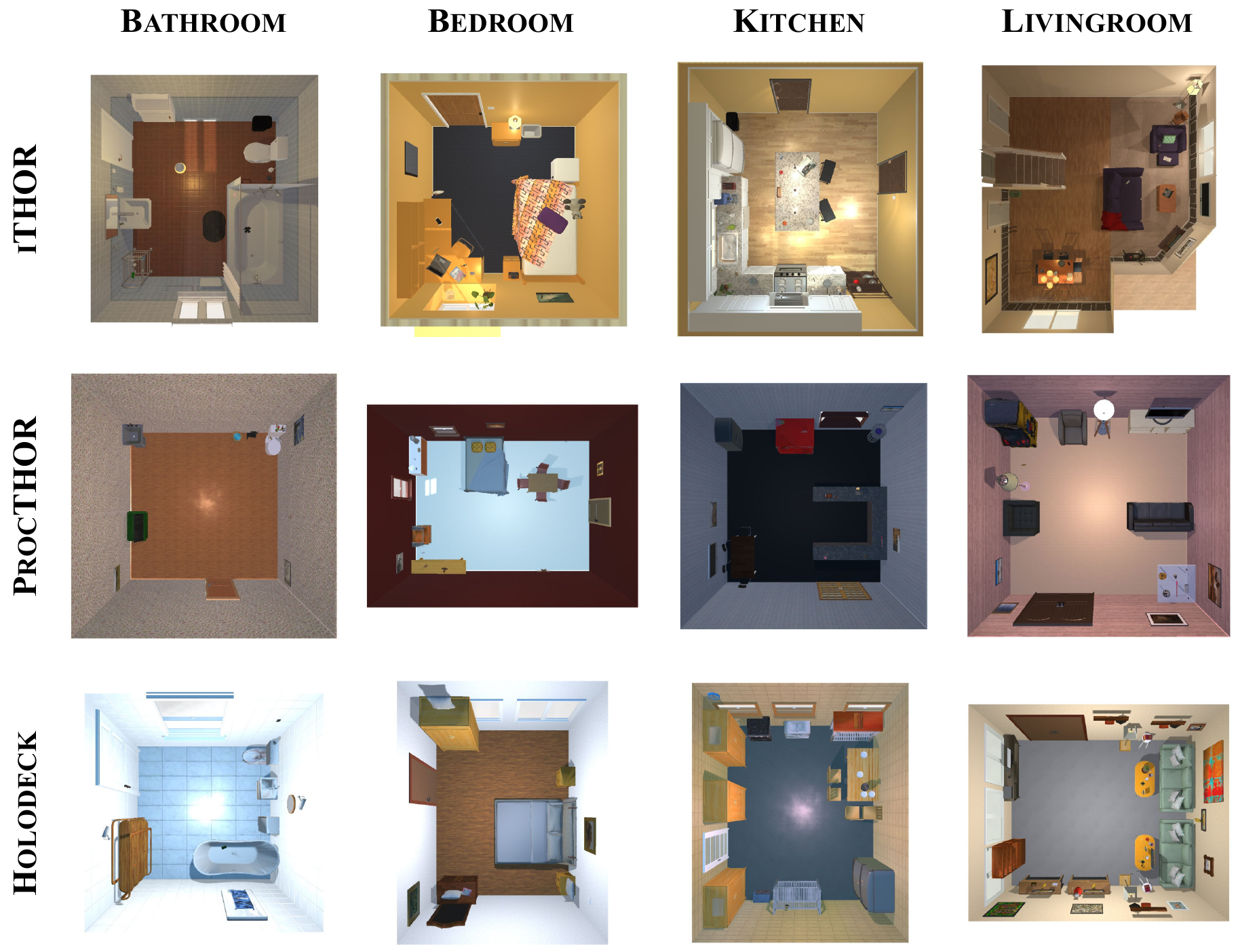}
    \caption{Qualitative examples on four types of residential scenes from iTHOR, \procthor, and \holodeck.}
    \label{fig: system_compare}
    \vspace{-0.4cm}
\end{figure*}

In Figure \ref{fig: mit_scenes_add}, we showcase an additional 20 scenes generated by \holodeck. These 20 scene types are chosen from the MIT dataset \cite{quattoni2009recognizing}, distinct from examples in the main paper. Figure \ref{fig: layout} presents a comparative analysis of layouts created by five methods. Figure \ref{fig: system_compare} offers a visual comparison of residential scenes from iTHOR, \procthor, and \holodeck, highlighting the differences and capabilities of each system.

\section{\noveltythor}
\noveltythor comprises human-designed scenes crafted to challenge embodied agents in unique and diverse environments with a wide array of assets from Objaverse.

To integrate Objaverse assets into Unity, we developed tools that run a conversion pipeline on various operating systems, including macOS and Windows. This flexibility also enables the inclusion of assets other than those found in Objaverse. We designed a user-friendly interface for our artists and designers, facilitating asynchronous asset integration while optimizing storage efficiency.

\begin{figure*}[!b]
\centering
\includegraphics[width=16cm]{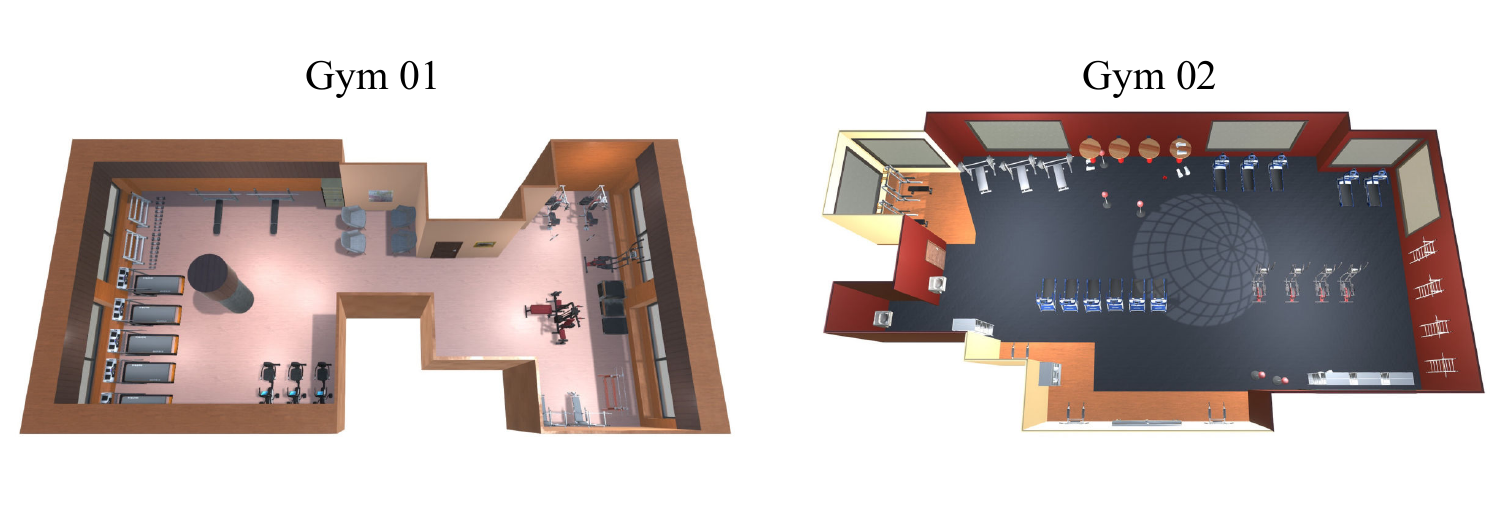}
    \caption{Top-down view of \noveltythor. Each scene type has two instances.}
    \label{fig: noveltythor}
    \vspace{-0.4cm}
\end{figure*}

\begin{figure*}[!t]
\centering
\includegraphics[width=16cm]{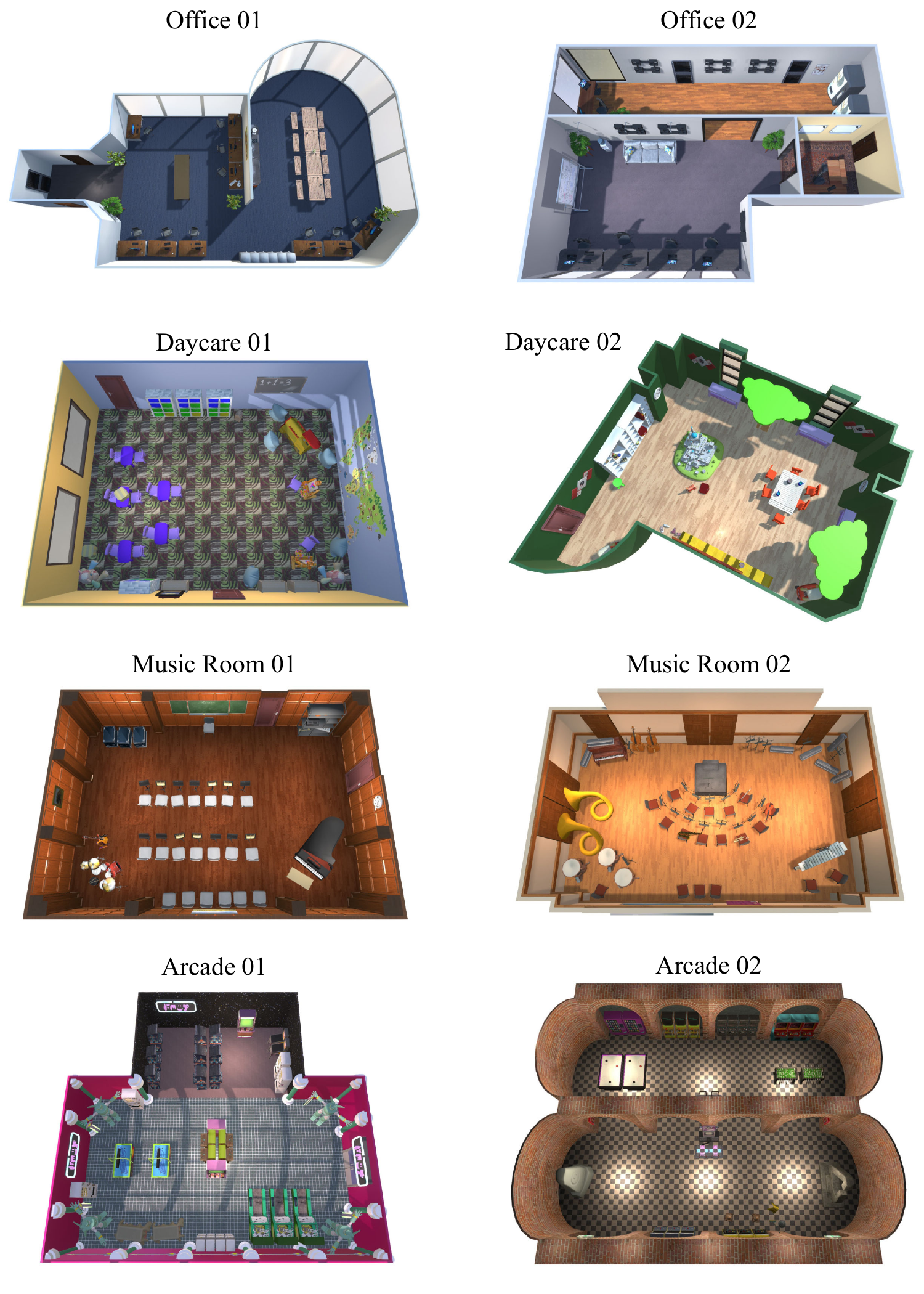}
    \caption{Top-down view of \noveltythor (continued).}
    \label{fig: noveltythor_continued}
    \vspace{-0.4cm}
\end{figure*}

The critical step of this process is the generation of Unity templates (prefabs) for the assets and their associated resources, leading to the creation of the scenes discussed in this paper. Figures \ref{fig: noveltythor} and \ref{fig: noveltythor_continued} showcase top-down views of the 10 \noveltythor scenes, spanning five categories. 

\begin{figure}[!t]
\centering
\includegraphics[width=8.3cm]{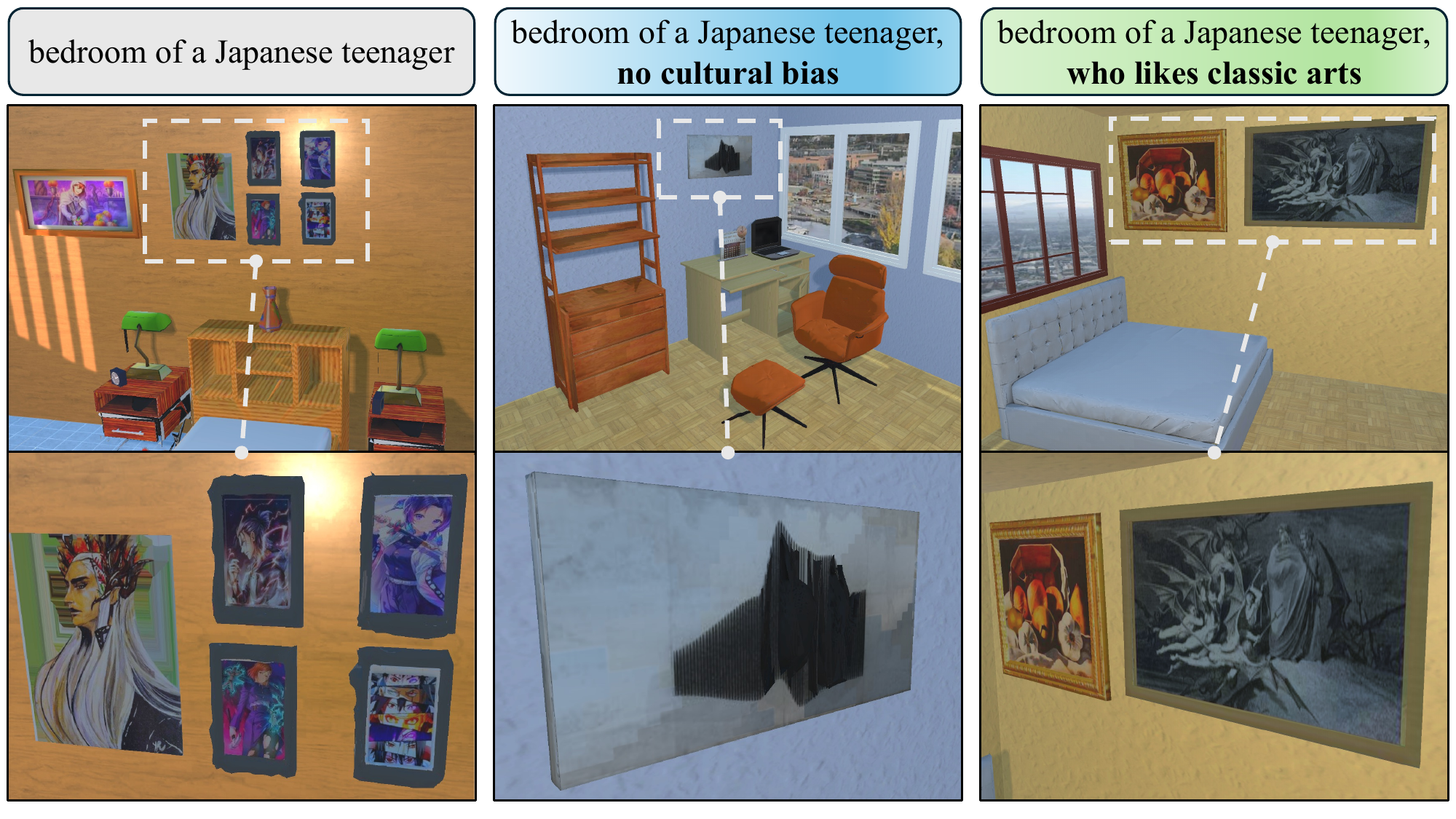}
    \caption{We can address the cultural bias of GPT-4 by prompting.}
    \label{fig: mitigate bias}
    \vspace{-0.4cm}
\end{figure}

\section{Cultural Bias}
Cultural biases in \holodeck generation can stem from biases in the LLM and the 3D asset retrieval component. For example, in Figure \ref{fig: mitigate bias} (left), when the prompt contains culturally specific terms such as ``Japanese'', the generated scene may disproportionately feature prototypical objects like Manga posters. One mitigation strategy is to adjust the prompts, e.g., users can control the generation by simply adding a suffix like ``no cultural bias'' or making the prompt more detailed. This strategy is unlikely to fully remove bias, but these qualitative results suggest it can significantly help.

\begin{figure*}[!b]
    \centering
    \begin{center}
    
    \begin{tcolorbox} [top=2pt,bottom=2pt, width=\linewidth, boxrule=1pt]
    {\footnotesize {\fontfamily{zi4}\selectfont
    \textbf{Floor plan Prompt:}
    You are an experienced room designer. Please assist me in crafting a floor plan. Each room is a rectangle. You need to define the four coordinates and specify an appropriate design scheme, including each room's color, material, and texture.
    Assume the wall thickness is zero. Please ensure that all rooms are connected, not overlapped, and do not contain each other. The output should be in the following format: room name | floor material | wall material | vertices (coordinates). Note: the units for the coordinates are meters. \\
    For example: \\
    living room | maple hardwood, matte | light grey drywall, smooth | [(0, 0), (0, 8), (5, 8), (5, 0)]\\
    kitchen | white hex tile, glossy | light grey drywall, smooth | [(5, 0), (5, 5), (8, 5), (8, 0)] \\
    
    Here are some guidelines for you: \\
    1. A room's size range (length or width) is 3m to 8m. The maximum area of a room is 48 m$^2$. Please provide a floor plan within this range and ensure the room is not too small or too large. \\
    2. It is okay to have one room in the floor plan if you think it is reasonable. \\
    3. The room name should be unique. \\
    
    Now, I need a design for \{input\}. \\
    Additional requirements: \{additional\_requirements\}. \\
    
    Your response should be direct and without additional text at the beginning or end.
    }
    \par}
    \end{tcolorbox}

    \begin{tcolorbox} [top=2pt,bottom=2pt, width=\linewidth, boxrule=1pt]
    {\footnotesize {\fontfamily{zi4}\selectfont
    \textbf{Wall Height Prompt:}
    I am now designing \{input\}. Please help me decide the wall height in meters.
    Answer with a number, for example, 3.0. Do not add additional text at the beginning or in the end.
    }
    \par}
    \end{tcolorbox}

    \begin{tcolorbox} [top=2pt,bottom=2pt, width=\linewidth, boxrule=1pt]
    {\footnotesize {\fontfamily{zi4}\selectfont
    \textbf{Doorway Prompt:}
    I need assistance in designing the connections between rooms. The connections could be of three types: doorframe (no door installed), doorway (with a door), or open (no wall separating rooms). The sizes available for doorframes and doorways are single (1m wide) and double (2m wide). \\
    
    Ensure that the door style complements the design of the room. The output format should be: room 1 | room 2 | connection type | size | door style.
    For example: \\
    exterior | living room | doorway | double | dark brown metal door \\
    living room | kitchen | open | N/A | N/A \\
    living room | bedroom | doorway | single | wooden door with white frames \\
    
    The design under consideration is \{input\}, which includes these rooms: \{rooms\}. \\
    The length, width, and height of each room in meters are:
    \{room\_sizes\} \\
    Certain pairs of rooms share a wall: \{room\_pairs\}. There must be a door to the exterior. \\
    Adhere to these additional requirements \{additional\_requirements\}. \\
    
    Provide your response succinctly, without additional text at the beginning or end.
    }
    \par}
    \end{tcolorbox}

    \begin{tcolorbox} [top=2pt,bottom=2pt, width=\linewidth, boxrule=1pt]
    {\footnotesize {\fontfamily{zi4}\selectfont
    \textbf{Window Prompt:}
    Guide me in designing the windows for each room. The window types are: fixed, hung, and slider. \\
    The available sizes (width x height in cm) are: \\
    fixed: (92, 120), (150, 92), (150, 120), (150, 180), (240, 120), (240, 180) \\
    hung: (87, 160), (96, 91), (120, 160), (130, 67), (130, 87), (130, 130) \\
    slider: (91, 92), (120, 61), (120, 91), (120, 120), (150, 92), (150, 120) \\
    
    Your task is to determine the appropriate type, size, and quantity of windows for each room, bearing in mind the room's design, dimensions, and function.
    
    Please format your suggestions as follows: room | wall direction | window type | size | quantity | window base height (cm from floor). For example:
    living room | west | fixed | (130, 130) | 1 | 50 \\
    
    I am now designing \{input\}. The wall height is \{wall\_height\} cm. \\
    The walls available for window installation (direction, width in cm) in each room are:
    \{walls\} \\
    Please note: It is not mandatory to install windows on every available wall. Within the same room, all windows must be the same type and size.
    Also, adhere to these additional requirements: \{additional\_requirements\}. \\
    
    Provide a concise response, omitting any additional text at the beginning or end. 
    }
    \par}
    \end{tcolorbox}
    
    \end{center}
    \caption{Prompt templates for Floor Module, Wall Module, Doorway Module, and Window Module.}
    \label{fig:prompt-1}
    \vspace{-.3cm}
\end{figure*}
\begin{figure*}[!t]
    \centering
    \begin{center}
    
    \begin{tcolorbox} [top=2pt,bottom=2pt, width=\linewidth, boxrule=1pt]
    {\footnotesize {\fontfamily{zi4}\selectfont
    \textbf{Object Selection Prompt:}
    You are an experienced room designer, please assist me in selecting large *floor*/*wall* objects and small objects on top of them to furnish the room. You need to select appropriate objects to satisfy the customer's requirements.
    You must provide a description and desired size for each object since I will use it to retrieve objects. If multiple identical items are to be placed in the room, please indicate the quantity and variance type (same or varied).
    Present your recommendations in JSON format:

    \{ 
        object\_name:\{ \\
            "description": a short sentence describing the object, \\
            "location": "floor" or "wall", \\
            "size": the desired size of the object, in the format of a list of three numbers, [length, width, height] in centimeters, \\
            "quantity": the number of objects (int), \\
            "variance\_type": "same" or "varied", \\
            "objects\_on\_top": a list of small children objects (can be empty) which are placed *on top of* this object. For each child object, you only need to provide the object\ name, quantity and variance type. For example, \{"object\_name": "book", "quantity": 2, "variance\_type": "varied"\}
        \}
    \} \\
    
    *ONE-SHOT EXAMPLE* \\
    
    Currently, the design in progress is *INPUT*, and we are working on the *ROOM\_TYPE* with the size of ROOM\_SIZE.
    Please also consider the following additional requirements: REQUIREMENTS.
    
    Here are some guidelines for you: \\
    1. Provide reasonable type/style/quantity of objects for each room based on the room size to make the room not too crowded or empty. \\
    2. Do not provide rug/mat, windows, doors, curtains, and ceiling objects which have been installed for each room. \\
    3. I want at least 10 types of large objects and more types of small objects on top of the large objects to make the room look more vivid. \\
    
    Please first use natural language to explain your high-level design strategy for *ROOM\_TYPE*, and then follow the desired JSON format *strictly* (do not add any additional text at the beginning or end).
    }
    \par}
    \end{tcolorbox}

    \begin{tcolorbox} [top=2pt,bottom=2pt, width=\linewidth, boxrule=1pt]
    {\footnotesize {\fontfamily{zi4}\selectfont
    \textbf{Layout Design Prompt:}
    You are an experienced room designer.
    Please help me arrange objects in the room by assigning constraints to each object.
    Here are the constraints and their definitions: \\
    1. global constraint: \\
        1) edge: at the edge of the room, close to the wall, most of the objects are placed here. \\
        2) middle: not close to the edge of the room.
    
    2. distance constraint: \\
        1) near, object: near to the other object, but with some distanbce, 50cm < distance < 150cm. \\
        2) far, object: far away from the other object, distance >= 150cm.

    3. position constraint: \\
        1) in front of, object: in front of another object. \\
        2) side of, object: on the side (left or right) of another object.
    
    4. alignment constraint:
        1) center aligned, object: align the center of the object with the center of another object.
    
    5. Rotation constraint:
        1) face to, object: face to the center of another object. \\
    
    For each object, you must have one global constraint and you can select various numbers of constraints and any combinations of them and the output format must be:
    object | global constraint | constraint 1 | constraint 2 | ...
    
    For example:
    sofa-0 | edge \\
    coffee table-0 | middle | near, sofa-0 | in front of, sofa-0 | center aligned, sofa-0 | face to, sofa-0 \\
    tv stand-0 | edge | far, coffee table-0 | in front of, coffee table-0 | center aligned, coffee table-0 | face to, coffee table-0 \\
    
    Here are some guidelines for you: \\
    1. I will use your guideline to arrange the objects *iteratively*, so please start with an anchor object which doesn't depend on the other objects (with only one global constraint). \\
    2. Place the larger objects first. \\
    3. The latter objects could only depend on the former objects. \\
    4. The objects of the *same type* are usually *aligned*. \\
    5. I prefer objects to be placed at the edge (the most important constraint) of the room if possible which makes the room look more spacious. \\
    6. Chairs must be placed near to the table/desk and face to the table/desk. \\
    Now I want you to design \{room\_type\} and the room size is \{room\_size\}. \\
    Here are the objects that I want to place in the \{room\_type\}:
    \{objects\} \\
    Please first use natural language to explain your high-level design strategy, and then follow the desired format *strictly* (do not add any additional text at the beginning or end) to provide the constraints for each object.
    }
    \par}
    \end{tcolorbox}
    
    \end{center}
    \caption{Prompt templates for Object Selection Module and Layout Design Module.}
    \label{fig:prompt-2}
    \vspace{-.3cm}
\end{figure*}

\begin{figure*}[!t]
    \centering
    \begin{center}
    \begin{tcolorbox} [top=2pt,bottom=2pt, width=\linewidth, boxrule=1pt]
    {\footnotesize {\fontfamily{zi4}\selectfont
    \textbf{3D Asset annotation Prompt:}
   Please annotate this 3D asset with the following values (output valid JSON): \\
        "annotations": \{ \\
            "category": a category such as "chair", "table", "building", "person", "airplane", "car", "seashell", "fish", etc. Try to be more specific than "furniture", \\
            "synset": the synset of the object that is most closely related. This could be "cat.n.01", "glass.n.03", "bank.n.02", \\
            "width": approximate width in cm. For a human being, this could be "45", \\
            "length": approximate length in cm. For a human being, this could be "25", \\
            "height": approximate height in cm. For a human being, this could be "182", \\
            "volume": approximate volume in cm$^3$. For a human being, this could be "62000", \\
            "mass": approximate mass in kilogram. For a human being, this could be "72", \\
            "frontView": which of the views represents the front of the object (value should be an integer with the first image being 0). Note that the front view of an object, including furniture, tends to be the view that exhibits the highest degree of symmetry, \\
            "description": a description of the object (don't use the term "3D asset" here), \\
            "materials": a Python list of the materials that the object appears to be made of (roughly in order of most used material to least used), \\
            "onCeiling": whether this object can appear on the ceiling or not, return true or false with no explanations. This would be true for a ceiling fan but false for a chair, \\
            "onWall": whether this object can appear on the wall or not, return true or false with no explanations. This would be true for a painting but false for a table, \\
            "onFloor": whether this object can appear on the floor or not, return true or false with no explanations. This would be true for a piano but false for a curtain, \\
            "onObject": whether this object can appear on another object or not, return true or false with no explanations. This would be true for a laptop but not for a sofa
        \} \\ \\
Please output the JSON now.
}
    \par}
    \end{tcolorbox}

    \end{center}
    \caption{Prompt template for annotating 3D assets with GPT-4-V.}
    \label{fig:prompt-3}
\end{figure*}

\end{document}